\documentclass[10pt,journal,compsoc]{IEEEtran}
\ifCLASSOPTIONcompsoc
  \usepackage[nocompress]{cite}
\else
  \usepackage{cite}
\fi
\ifCLASSINFOpdf
   \usepackage[pdftex]{graphicx}
  \graphicspath{{figs/}}
\else
 \fi

\usepackage{amsmath}
\usepackage{epsfig}
\usepackage{epstopdf}

\usepackage{algorithm}
\usepackage[noend]{algpseudocode}

\usepackage{url}

\usepackage{enumitem}

\begin{document}

\title{HyperFace: A Deep Multi-task Learning Framework for Face Detection, Landmark Localization, Pose Estimation, and Gender Recognition}

\author{Rajeev~Ranjan,~\IEEEmembership{Member,~IEEE,}
        Vishal~M.~Patel,~\IEEEmembership{Senior Member,~IEEE,}
        and~Rama~Chellappa,~\IEEEmembership{Fellow,~IEEE}% <-this % stops a space
\IEEEcompsocitemizethanks{\IEEEcompsocthanksitem R. Ranjan and R. Chellappa are with the Department
of Electrical and Computer Engineering, University of Maryland, College Park,
MD, 20742.\protect\\
E-mail: \{rranjan1,rama\}@umiacs.umd.edu
\IEEEcompsocthanksitem V. M. Patel is with Rutgers University.}% <-this % stops an unwanted space
\thanks}

\markboth{IEEE Transactions on Pattern Analysis and Machine Intelligence ,~Vol.~XX, No.~XX, ~2016}%
{Shell \MakeLowercase{\textit{et al.}}: Bare Demo of IEEEtran.cls for Computer Society Journals}
\IEEEtitleabstractindextext{%
\begin{abstract}
We present an algorithm for simultaneous face detection, landmarks localization, pose estimation and gender recognition using deep convolutional neural networks (CNN). The proposed method called, HyperFace, fuses the intermediate layers of a deep CNN using a separate CNN followed by a multi-task learning algorithm that operates on the fused features. It exploits the synergy among the tasks which boosts up their individual performances. Additionally, we propose two variants of HyperFace: (1)~HyperFace-ResNet that builds on the ResNet-101 model and achieves significant improvement in performance, and   (2)~Fast-HyperFace that uses a high recall fast face detector for generating region proposals to improve the speed of the algorithm.  Extensive experiments show that the proposed models are able to capture both global and local information in faces and performs significantly better than many competitive algorithms for each of these four tasks.
\end{abstract}

}

\maketitle

\IEEEdisplaynontitleabstractindextext
\IEEEpeerreviewmaketitle

\IEEEraisesectionheading{\section{Introduction}\label{sec:introduction}}
\IEEEPARstart{D}{etection} and analysis of faces is a challenging problem in computer vision, and has been actively researched for applications such as face verification, face tracking, person identification, etc. Although recent methods based on deep Convolutional Neural Networks (CNN) have achieved remarkable results for the face detection task \cite{DDFD_ICMR2015}, \cite{FD_BTAS2015}, \cite{faceness_ICCV2015}, it is still difficult to obtain facial landmark locations, head pose estimates and gender information from face images containing  extreme poses, illumination and resolution variations. The tasks of face detection, landmark localization, pose estimation and gender classification have generally been solved as separate problems. Recently, it has been shown that learning correlated tasks simultaneously can boost the performance of individual tasks  \cite{FaceDPL} ,\cite{AFW_dataset_CVPR2012}, \cite{JointCascade_LI_ECCV2014}.

\begin{figure}[t]
\begin{center}
\includegraphics[width=1\linewidth]{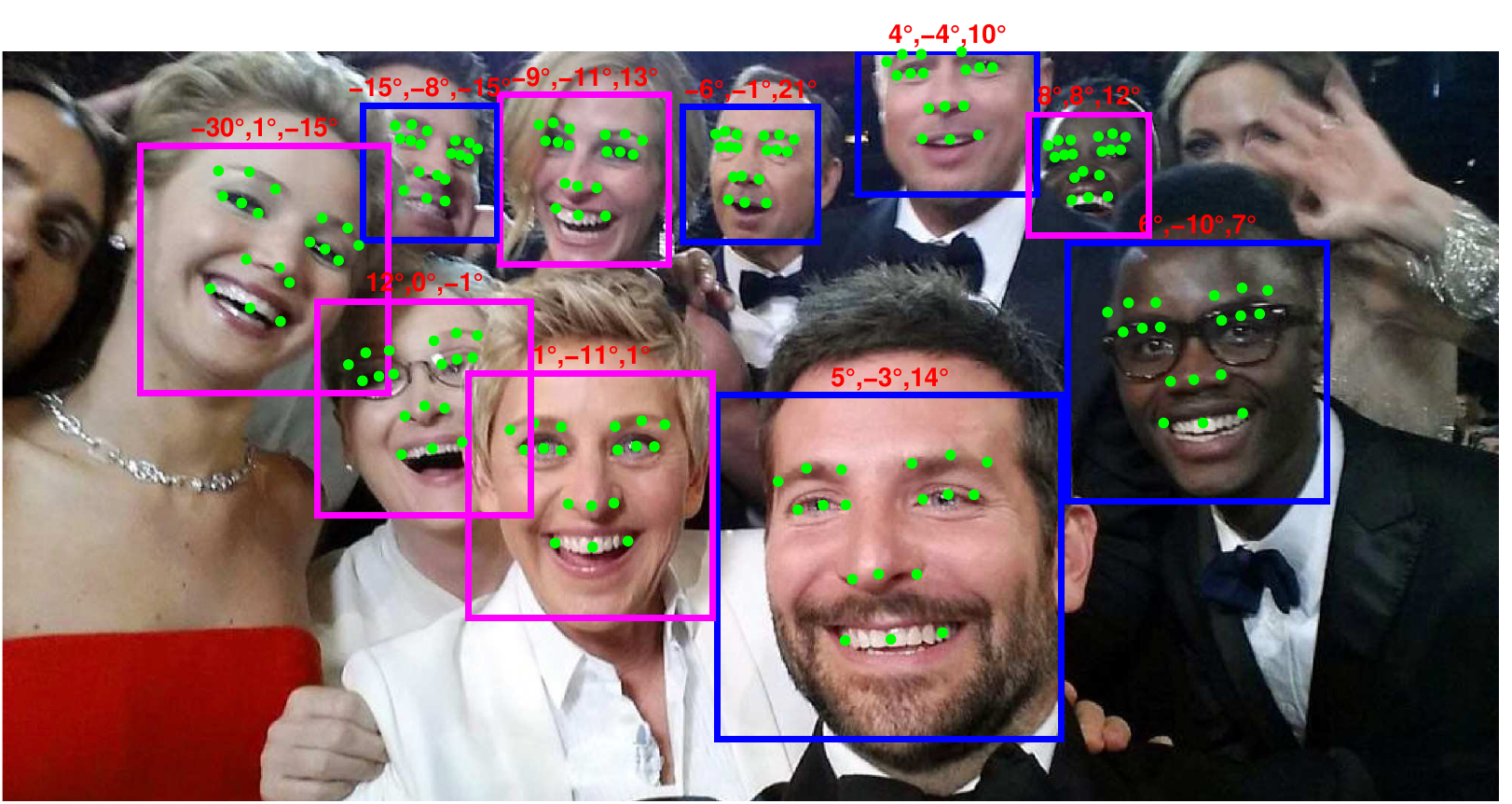}
\end{center}
   \caption{Our method can simultaneously detect the face, localize landmarks, estimate the pose and recognize the gender.  The blue boxes denote detected male faces, while pink boxes denote female faces. The green dots provide the landmark locations. Pose estimates for each face are shown on top of the boxes in the order of roll, pitch and yaw.}
\label{fig:overview}
\end{figure}

In this paper, we present a novel framework based on CNNs for simultaneous face detection, facial landmarks localization, head pose estimation and gender recognition from a given image (see Figure~\ref{fig:overview}). We design a CNN architecture to learn common features for these tasks and exploit the synergy among them. We exploit the fact that information contained in features is hierarchically distributed throughout the network as demonstrated in \cite{DBLP:journals/corr/ZeilerF13}.  Lower layers respond to edges and corners, and hence contain better localization properties. They are more suitable for learning landmarks localization and pose estimation tasks. On the other hand, deeper layers are class-specific and suitable for learning complex tasks such as face detection and gender recognition.  It is evident that we need to make use of all the intermediate layers of a deep CNN in order to train different tasks under consideration. We refer the set of intermediate layer features as \textit{hyperfeatures}. We borrow this term from \cite{Agarwal08multilevelimage} which uses it to denote a stack of local histograms for multilevel image coding.

\begin{figure*}[htp!]
 \centering
 \includegraphics[width=18cm]{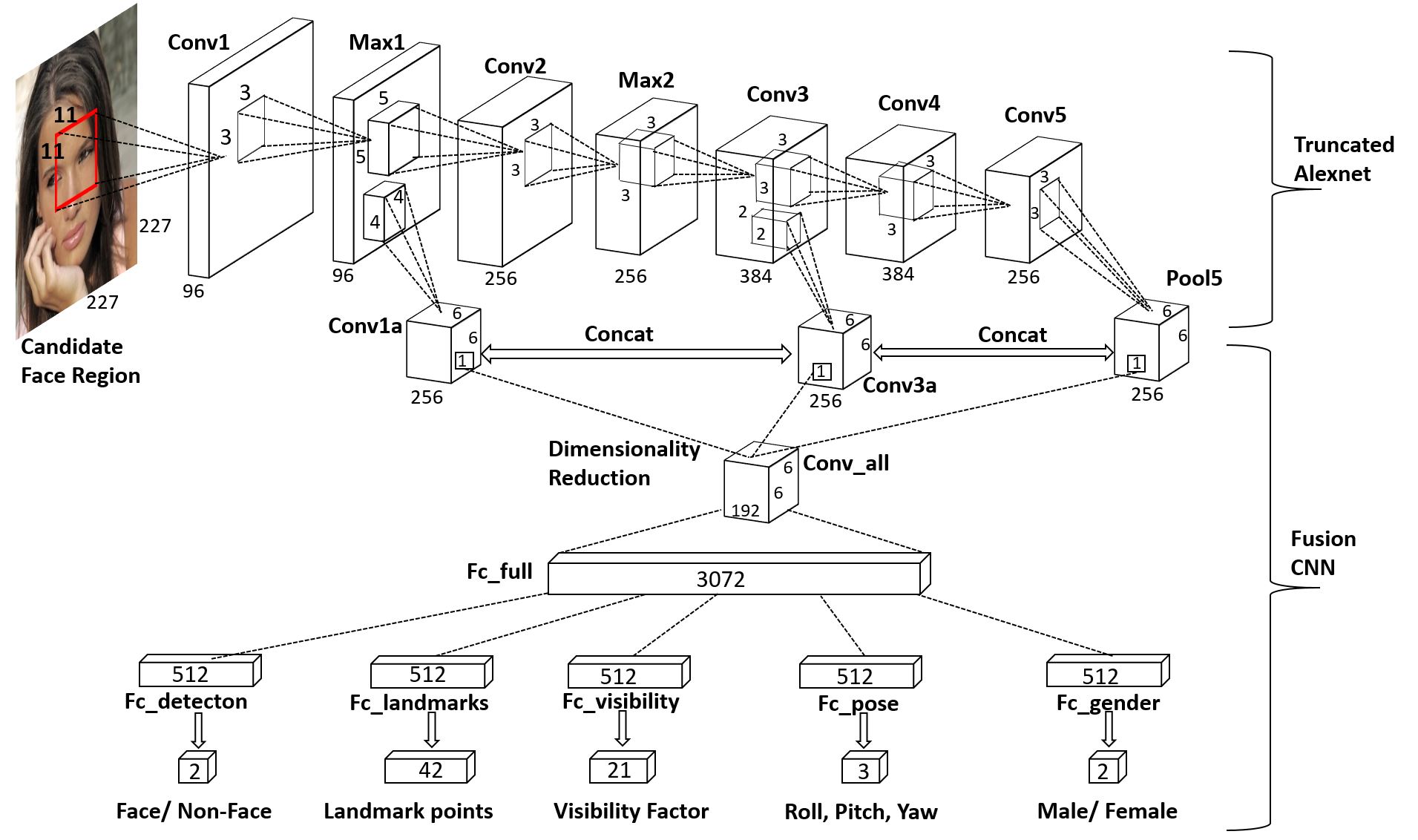}\\
 \caption{The architecture of the proposed HyperFace. The network is able to classify a given image region as face or non-face, estimate the head pose, locate face landmarks and recognize gender.}
\label{fig:model}
\end{figure*}

Since a CNN architecture contains multiple layers with hundreds of feature maps in each layer, the overall dimension of hyperfeatures is too large to be efficient for learning multiple tasks. Moreover, the hyperfeatures must be associated in a way that they efficiently encode the features common to the multiple tasks. This can be handled using feature fusion techniques. Features fusion aims to transform the features to a common subspace where they can be combined linearly or non-linearly. Recent advances in deep learning have shown that CNNs are capable of estimating an arbitrary complex function. Hence, we construct a separate fusion-CNN to fuse the hyperfeatures. In order to learn the tasks, we train them simultaneously using multiple loss functions. In this way, the features get better at understanding faces, which leads to improvements in the performances of individual tasks. The deep CNN combined with the fusion-CNN can be learned together in an end-to-end fashion. 

We also study the performance of face detection, landmarks localization, pose estimation and gender recognition tasks using off-the-shelf Region-based CNN (R-CNN~\cite{girshick14CVPR}) approach. Although R-CNN for face detection has been explored in DP2MFD~\cite{FD_BTAS2015}, we provide a comprehensive study of all these tasks based on R-CNN. Furthermore, we study the multitask approach without fusing the intermediate layers of CNN. Detailed experiments show that the multi-task learning method performs better than methods based on individual learning. Fusing the intermediate layer features provides additional performance boost.
  This paper makes the following contributions. 
\begin{enumerate}[noitemsep]
\item We propose two novel CNN architectures that perform face detection, landmarks localization, pose estimation and gender recognition by fusing the intermediate layers of the network. The first one called HyperFace is based on AlexNet~\cite{NIPS2012_4824} model, while the second one called HyperFace-ResNet (HF-ResNet) is based on ResNet-101~\cite{he2016deep} model.
\item We propose two post-processing methods: Iterative Region Proposals~(IRP) and Landmarks-based Non-Maximum Suppression~(L-NMS), which leverage the multi-task information obtained from the CNN to improve the overall performance.
\item We study the performance of R-CNN-based approaches for individual tasks and the multi-task approach without intermediate layer fusion.
\item We achieve significant improvement in performance on challenging unconstrained datasets for all of these four tasks.
\end{enumerate}

This paper is organized as follows. Section~\ref{sec:related} reviews related work.  Section~\ref{sec:hyperface} describes the proposed \textit{HyperFace} framework in detail. Section~\ref{sec:baselines} describes the implementation of R-CNN, Multitask\_Face and HF-ResNet approaches. Section~\ref{sec:res} provides the results of HyperFace and HF-ResNet along with R-CNN baselines on challenging datasets. Finally, Section~\ref{sec:discussion} concludes the paper with a brief summary and discussion.

\hfill 
 
\hfill 

\section{Related Work}
\label{sec:related}

{\bf{Multi-Task Learning:}} Multi-task learning (MTL) was first analyzed in detail by Caruana~\cite{caruana1998multitask}. Since then, several approaches have adopted MTL for solving different problems in computer vision. One of the earlier approaches for jointly addressing the tasks of face detection, pose estimation, and landmark localization was proposed in~\cite{AFW_dataset_CVPR2012} and later extended in~\cite{FaceDPL}.  This method is based on a mixture of trees with a shared pool of parts in the sense that every facial landmark is modeled as a part and uses global mixtures to capture the topological changes due to viewpoint variations.  A joint cascade-based method was recently proposed in~\cite{JointCascade_LI_ECCV2014} for simultaneously detecting faces and landmark points on a given image. This method yields improved detection performance by incorporating a face alignment step in the cascade structure. 

Multi-task learning using CNNs has also been studied recently. Eigen and Fergus~\cite{eigen2015predicting} proposed a multi-scale CNN for simultaneously predicting depth, surface normals and semantic labels from an image. They apply CNNs at three different scales where the output of the smaller scale network is fed as input to the larger one. UberNet~\cite{kokkinos2016ubernet} adopts a similar concept of simultaneously training low-, mid- and high-level vision tasks. It fuses all the intermediate layers of a CNN at three different scales of the image pyramid for multi-task training on diverse sets. Gkioxari et al.~\cite{DBLP:journals/corr/GkioxariHGM14} train a CNN for person pose estimation and action detection, using features only from the last layer. The use of MTL for face analysis is somewhat limited. Zhang et al.~\cite{TCDCN} used MTL-based CNN for facial landmark detection along with the tasks of discrete head yaw estimation, gender recognition, smile and glass detection. In their method, the predictions for all theses tasks were pooled from the same feature space. Instead, we strategically design the network architecture such that the tasks exploit low level as well as high level features of the network. We also jointly predict the task of face detection and landmark localization. These two tasks always go hand-in-hand and are used in most end-to-end face analysis systems.

{\bf{Feature Fusion:}} Fusing intermediate layers from CNN to bring both geometry and semantically rich features together has been used by quite a few methods. Hariharan et al.~\cite{BharathCVPR2015} proposed Hypercolumns to fuse pool2, conv4 and fc7 layers of AlexNet~\cite{NIPS2012_4824} for image segmentation. Yang and Ramanan~\cite{Yang_2015_ICCV} proposed DAG-CNNs, which extract features from multiple layers to reason about high, mid and low-level features for image classification. Sermanet et al.~\cite{Sermanet:2013:PDU:2514950.2516194} merge the 1st stage output of CNN to the classifier input after sub-sampling, for the application of pedestrian detection.

{\bf{Face detection:}}  Viola-Jones detector \cite{Viola_Jones} is a classic method which uses cascaded classifiers on Haar-like features to detect faces.  This method provides realtime face detection, but works best for full, frontal, and well lit faces.  Deformable Parts Model (DPM) \cite{DPM_PAMI2010}-based face detection methods have also been proposed in the literature where a face is essentially defined as a collection of parts \cite{AFW_dataset_CVPR2012}, \cite{HeadHunter_Mathias_ECCV2014}.   It has been shown that in unconstrained face detection, features like HOG or Haar wavelets do not capture the discriminative facial information at different illumination variations or poses.  To overcome these limitations, various deep CNN-based face detection methods have been proposed in the literature \cite{FD_BTAS2015}, \cite{CascadeCNN_CVPR2015}, \cite{faceness_ICCV2015}, \cite{DDFD_ICMR2015}, \cite{CCF_ICCV2015}.  These methods have produced state-of-the-art results on many challenging publicly available face detection datasets.  Some of the other recent face detection methods include NPDFaces \cite{NPDFace_PAMI2015}, PEP-Adapt \cite{PEP_Adapt_LI_ICCV2013}, and \cite{JointCascade_LI_ECCV2014}.

{\bf{Landmarks localization:}} Fiducial points extraction or landmarks localization is one of the most important steps in face recognition.  Several approaches have been proposed in the literature.  These include both regression-based \cite{DBLP:journals/ijcv/CaoWWS14}, \cite{Yan:2013:LCM:2586110.2586274}, \cite{XiongD13}, \cite{global2015xiong}, \cite{ERT}, \cite{GNDPM} and model-based \cite{Cootes:1995:ASM:206543.206547} ,\cite{Matthews:2004:AAM:993451.996344}, \cite{conf/eccv/LiangXWS08} methods. While the former learns the shape increment given a mean initial shape, the latter trains an appearance model to predict the keypoint locations. CNN-based landmark localization methods have also been proposed in recent years~\cite{Sun:2013:DCN:2514950.2516090}, \cite{TCDCN},\cite{DBLP:journals/corr/KumarRPC16} and have achieved remarkable performance. 

Although much work has been done for localizing landmarks for frontal faces, limited attention has been given to profile faces which occur more often in real world scenarios. Jourabloo and Liu recently proposed PIFA~\cite{Jourabloo_2015_ICCV} that estimates 3D landmarks for large pose face alignment by integrating a 3D point distribution model with a cascaded coupled-regressor. Similarly, 3DDFA~\cite{DBLP:journals/corr/ZhuLLSL15} fits a dense 3D model by estimating its parameters using a CNN. Zhu et al.~\cite{zhu2016unconstrained} proposed a cascaded compositional learning approach that combines shape prediction from multiple domain specific regressors.

{\bf{Pose estimation:}} The task of head pose estimation is to infer the orientation of person's head relative to the camera view. It is useful in face verification for matching face similarity across different orientations. Non-linear manifold-based methods have been proposed in \cite{4270305}, \cite{1530062}, \cite{1047456} to classify face images based on pose.  A survey of various head pose estimation methods is provided in \cite{4497208}.

{\bf{Gender recognition:}}  Previous works on gender recognition have focused on finding good discriminative features for classification.  Most previous methods use one or more combination of features such as LBP, SURF, HOG or SIFT.  In recent years, attribute-based methods for face recognition have gained a lot of traction.  Binary classifiers were used in \cite{facetracer} for each attribute such as male, long hair, white etc.  Separate features were computed for different attributes and they were  used to train individual SVMs for each attribute.  CNN-based methods have also been proposed for learning attribute-based representations in~\cite{CelebA}, \cite{PANDA}.
%I wish you the best of success.

\hfill 
 
\hfill 

\section{HyperFace}
\label{sec:hyperface}
We propose a single CNN model for simultaneous face detection, landmark localization, pose estimation and gender classification.  The network architecture is deep in both vertical and horizontal directions, i.e., it has both top-down and lateral connections, as shown in Figure~\ref{fig:model}. In this section, we provide a brief overview of the system and then discuss the different components in detail.

The proposed algorithm called \textit{HyperFace} consists
of three modules. The first one generates  class independent region-proposals from the given image and scales them to $227\times227$ pixels. The second module is a CNN which takes in the resized candidate regions and classifies them as face or non-face. If a region gets classified as a face, the network additionally provides facial landmarks locations, estimated head pose and gender information. The third module is a post-processing step which involves Iterative Region Proposals~(IRP) and Landmarks-based Non-Maximum Suppression~(L-NMS) to boost the face detection score and improve the performance of individual tasks.

\subsection{HyperFace Architecture}

We start with Alexnet~\cite{NIPS2012_4824} for image classification. The network consists of five convolutional layers along with three fully connected layers.  We initialize the network with the weights of R-CNN\_Face network trained for face detection task as described in Section~\ref{sec:baselines}. All the fully connected layers are removed as they encode image-classification specific information, which is not needed for pose estimation and landmarks extraction. We exploit the following two observations to create our network. 1) The features in CNN are distributed hierarchically in the network. While the lower layer features are effective for landmarks localization and pose estimation, the higher layer features are suitable for more complex tasks such as detection or classification\cite{DBLP:journals/corr/ZeilerF13}. 2) Learning multiple correlated tasks simultaneously builds a synergy and improves the performance of individual tasks as shown in \cite{JointCascade_LI_ECCV2014,TCDCN}. Hence, in order to simultaneously learn face detection, landmarks, pose and gender, we need to fuse the features from the intermediate layers of the network (hyperfeatures), and learn multiple tasks on top of it. Since the adjacent layers are highly correlated, we do not consider all the intermediate layers for fusion. 

We fuse the $max_{1}$, $conv_{3}$ and $pool_{5}$ layers of Alexnet, using a separate network. A naive way for fusion is directly concatenating the features. Since the feature maps for these layers have different dimensions $27\times 27\times96$, $13\times13\times384$, $6\times6\times 256$, respectively, they cannot be easily concatenated. We therefore add $conv_{1a}$ and $conv_{3a}$ convolutional layers to $pool_{1}$, $conv_{3}$ layers to obtain consistent feature maps of dimensions $6\times6\times256$ at the output. We then concatenate the output of these layers along with $pool_{5}$ to form a $6\times6\times768$ dimensional feature maps. The dimension is still quite high to train a multi-task framework. Hence, a $1\times1$ kernel  convolution layer ($conv_{all}$) is added to reduce the dimensions \cite{DBLP:journals/corr/SzegedyLJSRAEVR14} to $6\times6\times 192$. We add a fully connected layer ($fc_{all}$) to $conv_{all}$, which outputs a $3072$ dimensional feature vector. At this point, we split the network into five separate branches corresponding to the different tasks. We add $fc_{detection}$, $fc_{landmarks}$, $fc_{visibility}$, $fc_{pose}$ and $fc_{gender}$ fully connected layers, each of dimension 512, to $fc_{all}$. Finally, a fully connected layer is added to each of the branch to predict the individual task labels. After every convolution or a fully connected layer, we deploy the Rectified Linear Unit (ReLU). We do not include any pooling operation in the fusion network as it provides local invariance which is not desired for the face landmark localization task. Task-specific loss functions are then used to learn the weights of the network.

\subsection{Training}
\label{training}
We use the AFLW\cite{AFLW} dataset for training the HyperFace network. It contains $25,993$  faces  in  $21,997$  real-world  images with full pose, expression, ethnicity, age and gender variations. It provides annotations for $21$ landmark points per face, along with the face bounding-box, face pose (yaw, pitch and roll) and gender information. We randomly selected $1000$ images for testing, and used the rest for training the network. Different loss functions are used for training the tasks of face detection, landmark localization, pose estimation and gender classification. 

{\bf{Face Detection:}} We use the Selective Search~\cite{vandeSande:2011:SSS:2355573.2356474} algorithm in R-CNN~\cite{girshick14CVPR} to generate region proposals for faces in an image. A region having an Intersection over Union (IOU) overlap of more than $0.5$ with the ground truth bounding box is considered a positive sample ($l=1$). The candidate regions with IOU overlap less than $0.35$ are treated as negative instances ($l=0$). All the other regions are ignored. We use the softmax loss function given by (\ref{eq:det_loss}) for training the face detection task.
\begin{equation}
\label{eq:det_loss}
loss_{D} = -(1-l)\cdot \log(1-p)-l \cdot \log(p),
\end{equation}
where p is the probability that the candidate region is a face.
The probability values $p$ and $1-p$ are obtained from the last fully connected layer for the detection task.

{\bf{Landmarks Localization:}} We use $21$ point markups for face landmarks locations as provided in the AFLW\cite{AFLW} dataset. Since the faces have full pose variations, some of the landmark points are invisible. The dataset provides the annotations for the visible landmarks. We consider bounding-box regions with IOU overlap greater than $0.35$ with the ground truth for learning this task, while ignoring the rest. A region can be characterized by $\{x,y,w,h\}$ where $(x,y)$ are the co-ordinates of the center of the region and $w$,$h$ are the width and height of the region respectively. Each visible landmark point is shifted with respect to the region center $(x,y)$, and normalized by ($w,h$) as given by  (\ref{eq:lmark_norm})
\begin{equation}
\label{eq:lmark_norm}
(a_{i},b_{i}) = \left(\frac{x_{i}-x}{w},\frac{y_{i}-y}{h}\right).
\end{equation}
where ($x_{i},y_{i}$)'s are the given ground truth fiducial co-ordinates. The ($a_{i},b_{i}$)'s are treated as labels for training the landmark localization task using the Euclidean loss weighted by the visibility factor. The loss in predicting the landmark location is computed from  (\ref{eq:lmark_loss})
\begin{equation}
\label{eq:lmark_loss}
loss_{L} = \frac{1}{2N}\sum_{i=1}^{N}v_{i}((\hat{x_{i}}-a_{i})^2+((\hat{y_{i}}-b_{i})^2),
\end{equation}
where ($\hat{x_{i}},\hat{y_{i}}$) is the $i^{th}$ landmark location predicted by the network, relative to a given region, $N$ is the total number of landmark points ($21$ for AFLW\cite{AFLW}).  The visibility factor $v_{i}$ is $1$ if the $i^{th}$ landmark is visible in the candidate region, else it is $0$. This implies that there is no loss corresponding to invisible points and hence they do not take part during back-propagation.
%The labels for landmarks which are not visible are taken to be ($0,0$). 

{\bf{Learning Visibility:}} We also learn the visibility factor in order to test the presence of the predicted landmark. For a given region with overlap higher than $0.35$, we use a simple Euclidean loss to train the visibility as shown in  (\ref{eq:viz_loss})
\begin{equation}
\label{eq:viz_loss}
loss_{V} = \frac{1}{N}\sum_{i=1}^{N}\left(\hat{v_{i}}-v_{i}\right)^2,
\end{equation}
where $\hat{v_{i}}$ is the predicted visibility of $i^{th}$ landmark. The true visibility $v_{i}$ is $1$ if the $i^{th}$ landmark is visible in the candidate region, else it is $0$.

{\bf{Pose Estimation:}} We use the Euclidean loss to train the head pose estimates of roll ($p_{1}$), pitch ($p_{2}$) and yaw ($p_{3}$). We compute the loss for a candidate region having an overlap more than $0.5$ with the ground truth, from  (\ref{eq:pose_loss})
\begin{equation}
\label{eq:pose_loss}
loss_{P} = \frac{(\hat{p_{1}}-p_{1})^2+(\hat{p_{2}}-p_{2})^2+(\hat{p_{3}}-p_{3})^2}{3},
\end{equation}
where ($\hat{p_{1}},\hat{p_{2}},\hat{p_{3}}$) are the estimated pose labels.

{\bf{Gender Recognition:}} Predicting gender is a two class problem similar to face detection. For a candidate region with overlap of $0.5$ with the ground truth, we compute the softmax loss given in  (\ref{eq:gender_loss})
\begin{equation}
\label{eq:gender_loss}
loss_{G} = -(1-g)\cdot \log(1-p_{g})-g\cdot \log(p_{g}),
\end{equation}
where $g=0$ if the gender is male, or else $g=1$. Here, ($p_{0},p_{1}$) is the two dimensional probability vector computed from the network.

The total loss is computed as the weighted sum of the five individual losses as shown in  (\ref{eq:full_loss})
\begin{equation}
\label{eq:full_loss}
loss_{full} = \sum_{i=1}^{i=5}\lambda_{t_{i}}loss_{t_{i}},
\end{equation}
where $t_{i}$ is the $i^{th}$ element from the set of tasks $T = \{D, L, V, P, G\}$. The weight parameter $\lambda_{t_{i}}$ is decided based on the importance of the task in the overall loss. We choose ($\lambda_{D}=1,\lambda_{L}=5,\lambda_{V}=0.5,\lambda_{P}=5,\lambda_{G}=2$) for our experiments. Higher weights are assigned to landmark localization and pose estimation tasks as they need spatial accuracy.

\begin{figure}[htp!]
\begin{center}
\includegraphics[width=1\linewidth]{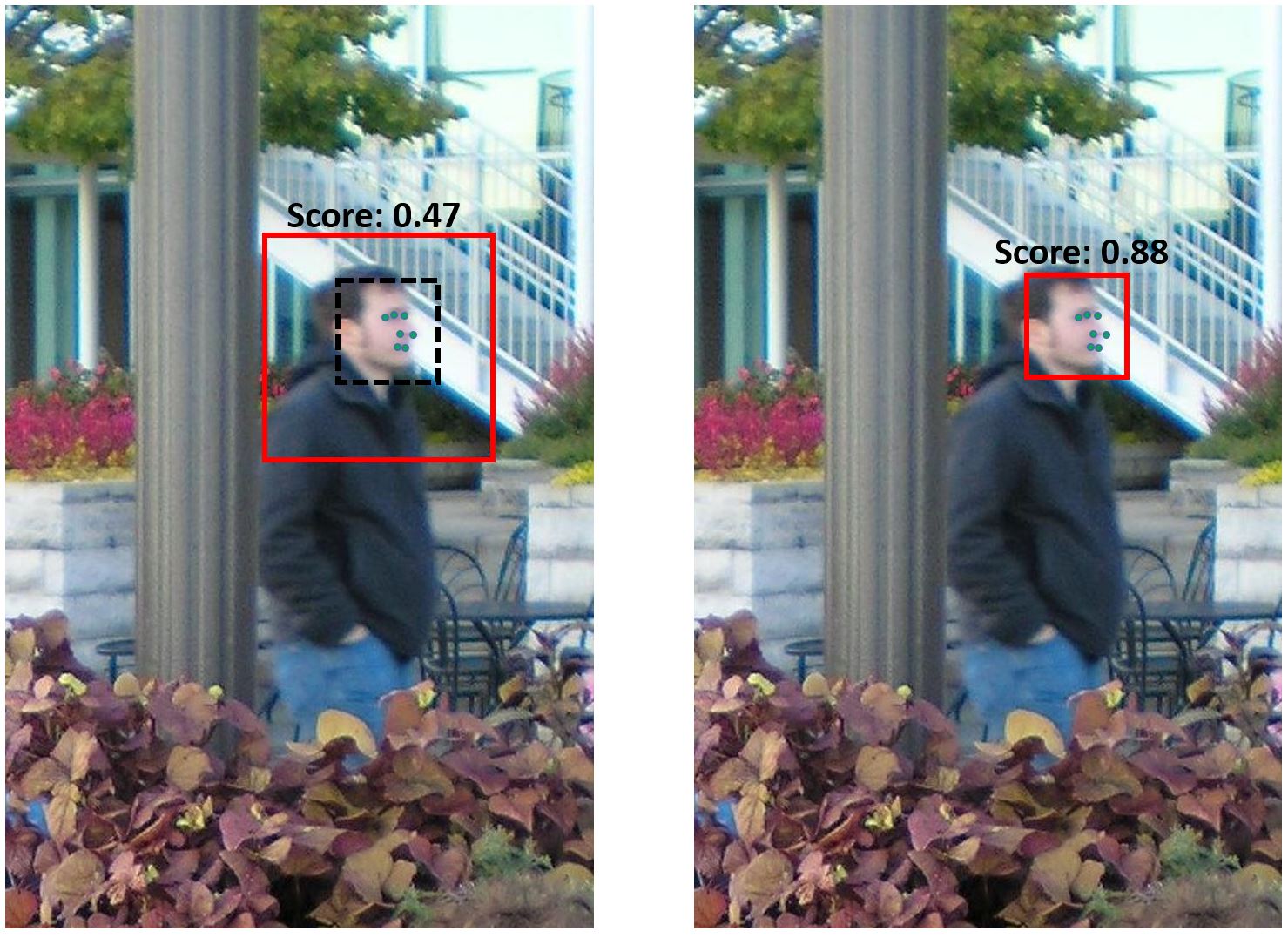}
\end{center}
   \caption{Candidate face region (red box on left) obtained using Selective Search gives a low score for face detection, while landmarks are correctly localized. We generate a new face region (red box on right) using the landmarks information and feed it through the network to increase the detection score.}
\label{fig:rrp}
\end{figure}

\subsection{Testing}
From a given test image, we first extract the candidate region proposals using\cite{vandeSande:2011:SSS:2355573.2356474}. For each region, we predict the task labels by a forward-pass through the HyperFace network. Only those regions, whose detection scores are above a certain threshold, are classified as face and processed for subsequent tasks. The predicted landmark points are scaled and shifted to the image co-ordinates using  (\ref{eq:lmark_unnorm})
\begin{equation}
\label{eq:lmark_unnorm}
(x_{i}, y_{i}) = (\hat{x_{i}}w + x, \hat{y_{i}} h + y),
\end{equation}
where ($\hat{x_{i}},\hat{y_{i}}$) are the predicted locations of the $i^{th}$ landmark from the network, and $\{x,y,w,h\}$ are the region parameters defined in  (\ref{eq:lmark_norm}).
Points obtained with predicted visibility less than a certain threshold are marked invisible. The pose labels obtained from the network are the estimated roll, pitch and yaw for the face region. The gender is assigned according to the label with maximum predicted probability.

There are two major issues while using proposal-based face detection. First, the proposals might not be able to capture small and difficult faces, hence reducing the overall recall of the system. Second, the proposal boxes might not be well localized with the actual face region. It is a common practice to use bounding-box regression~\cite{girshick14CVPR} as a post processing step to improve the localization of the detected face box. This adds an additional burden of training regressors to learn the transformation from the detected candidate box to the annotated face box. Moreover, the localization is still weak since the regressors are usually linear. Recently, Gidaris and Komodakis proposed LocNet~\cite{gidaris2016locnet} which tries to solve these limitations by refining the detection bounding box. Given a set of initial bounding box proposals, it generates new sets of bounding boxes that maximize the likelihood of each row and column within the box. It allows an accurate inference of bounding box under a simple probabilistic framework.  

Instead of using the probabilistic framework~\cite{gidaris2016locnet}, we solve the above mentioned issues in an iterative way using the predicted landmarks. The fact that we obtain landmark locations along with the detections, enables us to improve the post-processing step so that all the tasks benefit from it. We propose two novel methods: Iterative Region Proposals (IRP) and Landmarks-based Non-Maximum Suppression (L-NMS) to improve the performance. IRP improves the recall by generating more candidate proposals by using the predicted landmarks information from the initial set of region proposals. On the other hand, L-NMS improves the localization by re-adjusting the detected bounding boxes according to the predicted landmarks and performing NMS on top of them. No additional training is required for these methods.

{\bf{Iterative Region Proposals (IRP):}} We use a fast version of Selective Search\cite{vandeSande:2011:SSS:2355573.2356474} which extracts around $2000$ regions from an image. We call this version $Fast\_SS$. It is quite possible that some faces with poor illumination or small size fail to get captured by any candidate region with a high overlap. The network would fail to detect that face due to low score. In these situations, it is desirable to have a candidate box which precisely captures the face.  Hence, we generate a new candidate bounding box from the predicted landmark points using the FaceRectCalculator provided by \cite{AFLW}, and pass it again through the network. The new region, being more localized yields a higher detection score and improves the corresponding tasks output, thus increasing the recall. This procedure can be repeated (say $T$ time), so that boxes at a given step will be more localized to faces as compared to the previous step. From our experiments, we found that the localization component saturates in just one step ($T$ = $1$), which shows the strength of the predicted landmarks. The pseudo-code of IRP is presented in Algorithm~\ref{alg:irp}. The usefulness of IRP can be seen in Figure \ref{fig:rrp}, which shows a low-resolution face region cropped from the top-right image in Figure~\ref{fig:quantative_results}. 

\begin{algorithm}
\caption{Iterative Region Proposals}\label{alg:irp}
\begin{algorithmic}[1]

\State $\textbf{boxes}\gets selective\_search (\textbf{image})$
\State $\textbf{scores}\gets get\_hyperface\_scores (\textbf{boxes})$
\State $\textbf{detected\_boxes}\gets \textbf{boxes}(scores \geq threshold)$
\State $\textbf{new\_boxes}\gets \textbf{detected\_boxes}$
\For{\texttt{stage = 1~to~T}}
    \State $\textbf{fids}\gets get\_hyperface\_fiducials(\textbf{new\_boxes})$
    \State $\textbf{new\_boxes}\gets FaceRectCalculator(\textbf{fids})$
    \State $\textbf{deteced\_boxes}\gets [\textbf{deteced\_boxes} | \textbf{new\_boxes}]$   
\EndFor
\State \textbf{end}
\State $\textbf{final\_scores}\gets get\_hyperface\_scores (\textbf{detected\_boxes})$
\end{algorithmic}
\end{algorithm}

{\bf{Landmarks-based Non-Maximum Suppression (L-NMS):}} The traditional approach of non-maximum suppression involves selecting the top scoring region and discarding all the other regions with overlap more than a certain threshold. This method can fail in the following two scenarios: 1) If a region corresponding to the same detected face has less overlap with the highest scoring region, it can be detected as a separate face. 2) The highest scoring region might not always be localized well for the face, which can create some discrepancy if two faces are close together. To overcome these issues, we perform NMS on a new region whose bounding box is defined by the boundary co-ordinates as $[\min_{i} x_{i},\min_{i}y_{i},\max_{i}x_{i},\max_{i}y_{i}]$ of the landmarks for the given region. In this way, the candidate regions would get close to each other, thus decreasing the ambiguity of the overlap and improving the localization. 

\begin{algorithm}
\caption{Landmarks-based NMS}\label{alg:lnms}
\begin{algorithmic}[1]

\State Get \textbf{detected\_boxes} from Algorithm~\ref{alg:irp}
\State $\textbf{fids}\gets get\_hyperface\_fiducials(\textbf{detected\_boxes})$
\State $\textbf{precise\_boxes}\gets [min_{x},min_{y},max_{x},max_{y}](\textbf{fids})$
\State $\textbf{faces}\gets nms(\textbf{precise\_boxes},overlap)$
\For{\textbf{each} \texttt{face in \textbf{faces}}}
    \State $\textbf{top-k\_boxes}\gets$ Get top-k scoring boxes
    \State $\textbf{final\_fids}\gets median(fids(\textbf{top-k\_boxes}))$
    \State $\textbf{final\_pose}\gets median(pose(\textbf{top-k\_boxes}))$
    \State $\textbf{final\_gender}\gets median(gender(\textbf{top-k\_boxes}))$
     \State $\textbf{final\_visibility}\gets median(visibility(\textbf{top-k\_boxes}))$
     \State $\textbf{final\_bounding\_box}\gets FaceRectCalculator(\textbf{final\_fids})$
\EndFor
\State \textbf{end}

\end{algorithmic}
\end{algorithm}

 We apply landmarks-based NMS to keep the top-$k$ boxes, based on the detection scores. The detected face corresponds to the region with maximum score. The landmark points, pose estimates and gender classification scores are decided by the median of the top $k$ boxes obtained. Hence, the predictions do not rely only on one face region, but considers the votes from top-$k$ regions for generating the final output. From our experiments, we found that the best results are obtained with the value of $k$ being $5$. The pseudo-code for L-NMS is given in Algorithm~\ref{alg:lnms}.

\hfill

\hfill
\begin{figure*}[htp!]
 \centering
\includegraphics[width=2.5cm, height=9.4cm]{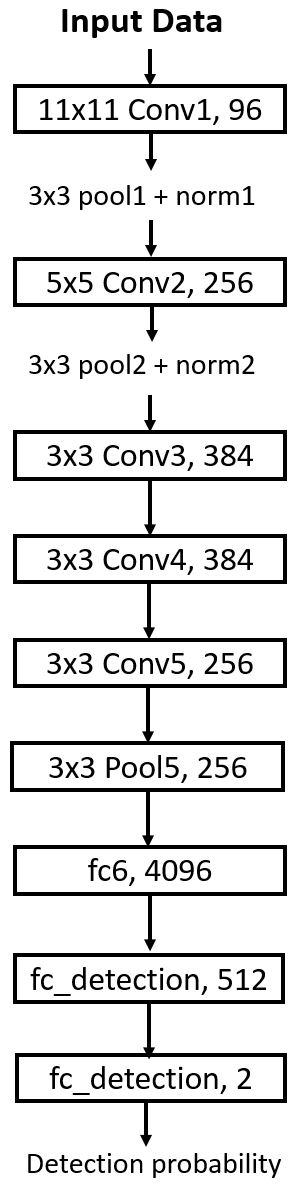}\hskip5pt\includegraphics[width=5.5cm, height=9.4cm]{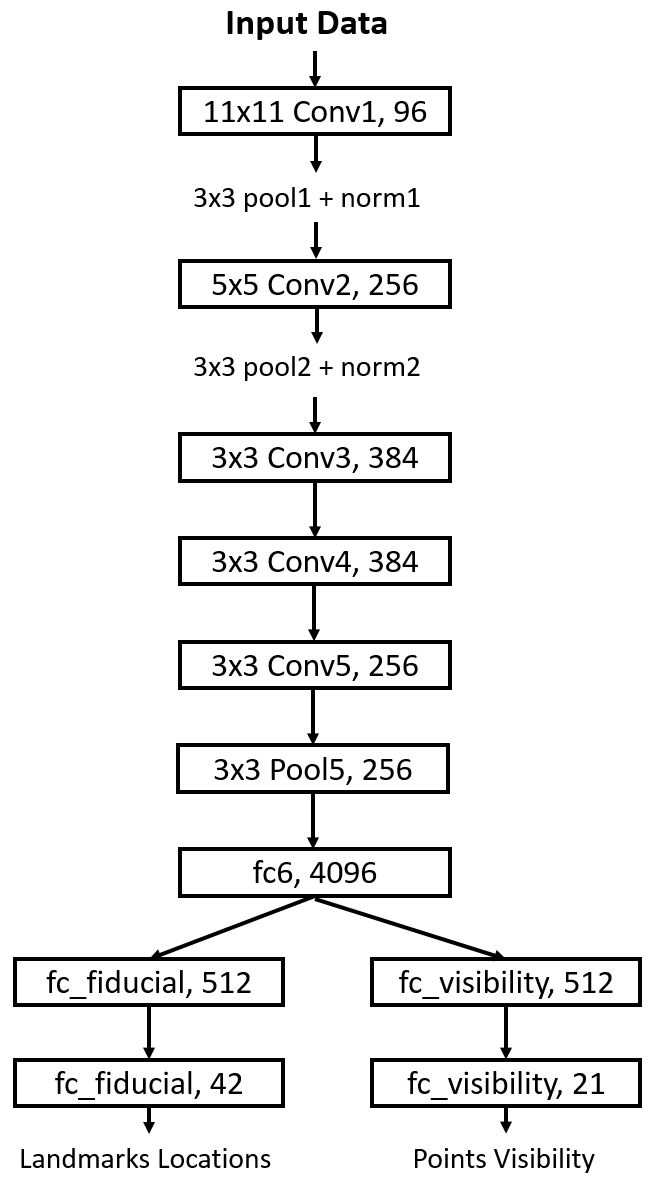}\hskip5pt\includegraphics[width=2.5cm, height=9.4cm]{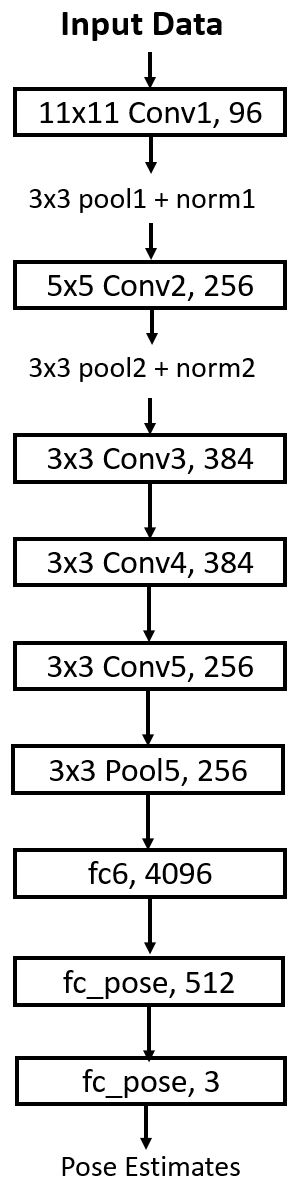}\hskip5pt\includegraphics[width=2.5cm, height=9.4cm]{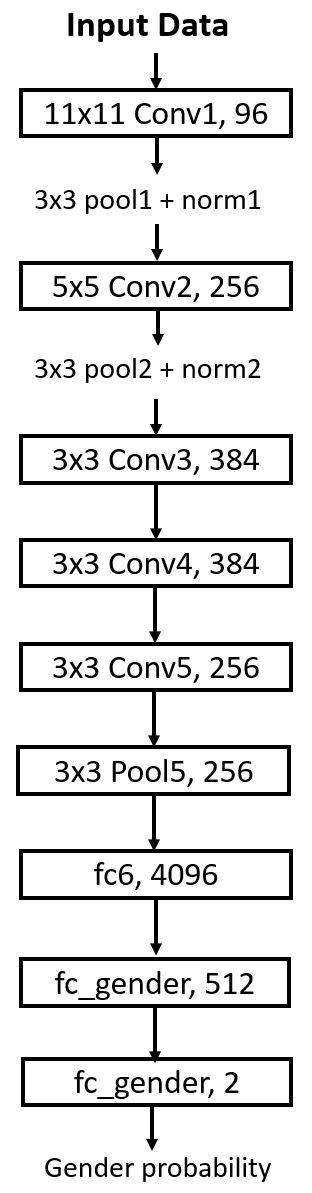}\\
(a)\hskip110pt(b)\hskip110pt(c)\hskip70pt(d)\\
\caption{R-CNN-based network architectures for (a) Face Detection (R-CNN\_Face), (b) Landmark Localization (R-CNN\_Fiducial), (c) Pose Estimation (R-CNN\_Pose), and (d) Gender Recognition (R-CNN\_Gender). The numbers on the left denote the kernel size and the numbers on the right denote the cardinality of feature maps for a given layer.}
\label{fig:rcnns}
\end{figure*}

\section{Network Architectures}
\label{sec:baselines}
To emphasize the importance of multitask approach and fusion of the intermediate layers of CNN, we study the performance of simpler CNNs devoid of such features. We evaluate four R-CNN-based models, one for each task of face detection, landmark localization, pose estimation and gender recognition. We also build a separate Multitask\_Face model which performs multitask learning just like HyperFace, but does not fuse the information from the intermediate layers. These models are described as follows:

{\bf{R-CNN\_Face:}} This model is used for face detection task. The network architecture is shown in Figure~\ref{fig:rcnns}(a). For training R-CNN\_Face, we use the region proposals from AFLW\cite{AFLW} training set, each associated with a face label based on the overlap with the ground truth. The loss is computed as per \eqref{eq:det_loss}. The model parameters are initialized using the Alexnet\cite{NIPS2012_4824} weights trained on the Imagenet dataset~\cite{deng2009imagenet}. Once trained, the learned parameters from this network are used to initialize other models including Multitask\_Face and HyperFace as the standard Imagenet initialization doesn't converge well. We also perform a linear bounding box regression to localize the face co-ordinates.

{\bf{R-CNN\_Fiducial:}} This model is used for locating the facial landmarks. The network architecture is shown in Figure~\ref{fig:rcnns}(b). It simultaneously learns the visibility of the points to account for the invisible points at test time, and thus can be used as a standalone fiducial extractor. The loss functions for landmarks localization and visibility of points are computed using \eqref{eq:lmark_loss} and \eqref{eq:viz_loss}, respectively. Only region proposals which have an overlap$>0.5$ with the ground truth bounding box are used for training. The model parameters are initialized from R-CNN\_Face.

\begin{figure}[htp!]
\centering
\includegraphics[width=8.5cm, height=10.0cm]{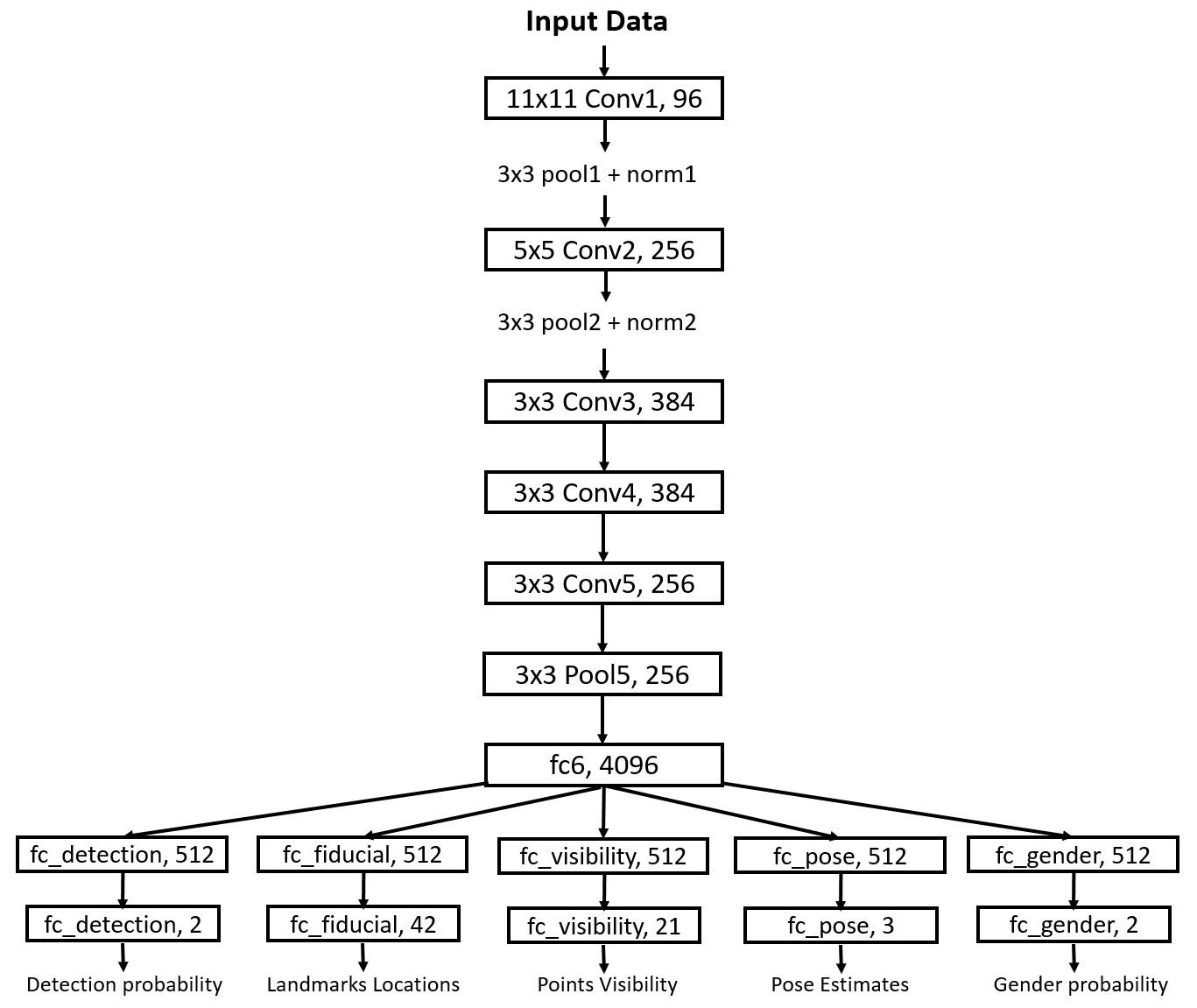}
\caption{Network Architecture of Multitask\_Face. The numbers on the left denote the kernel size and the numbers on the right denote the cardinality of feature maps for a given layer.}
\label{fig:multitask_face}
\end{figure}

\begin{figure*}[htp!]
	\centering
	\includegraphics[width=18cm]{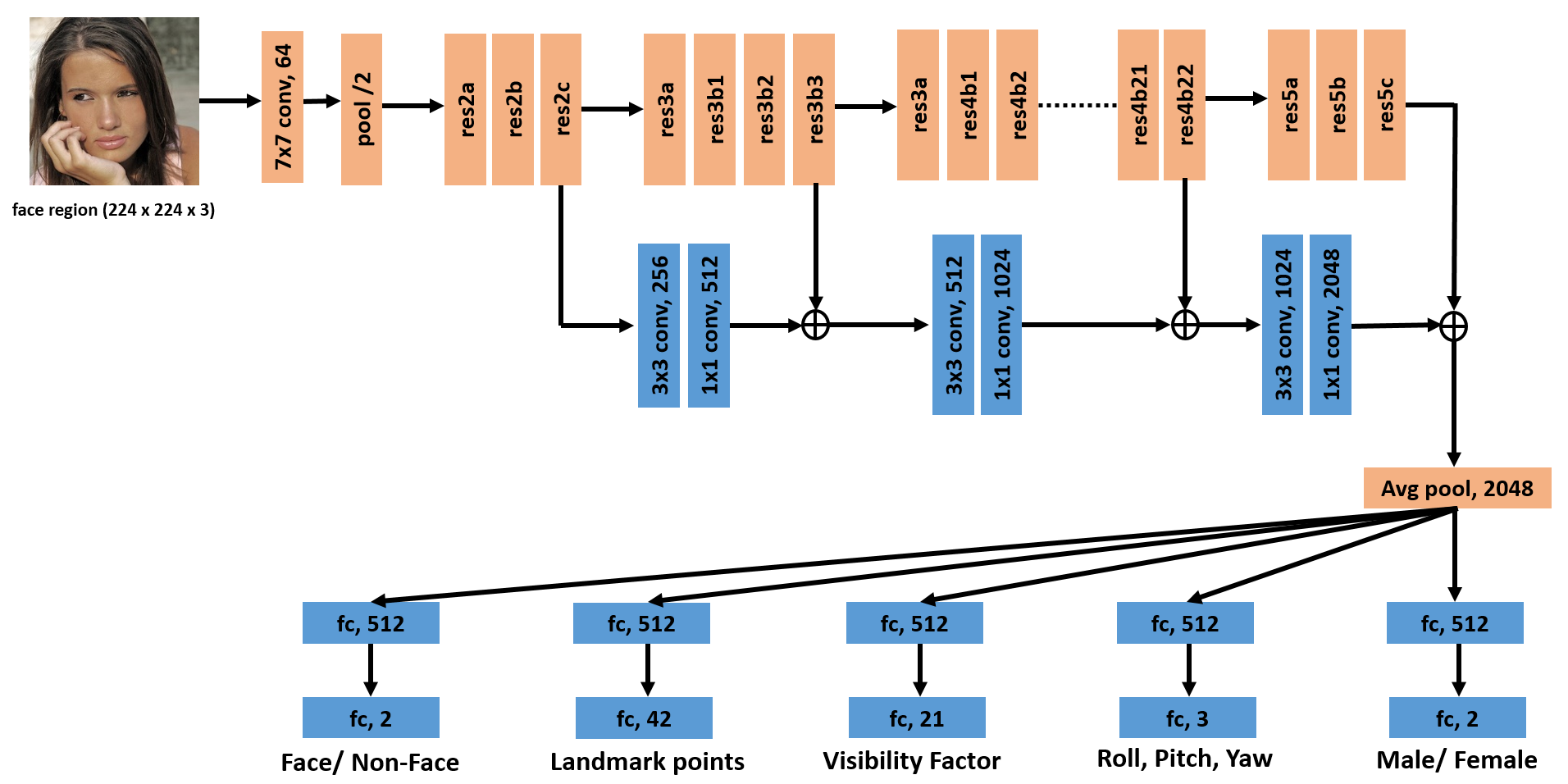}\\
	\caption{The architecture of the proposed HyperFace-Resnet (HF-ResNet). ResNet-101 model is used as the backbone network, represented in color orange. The new layers added are represented in color blue. The network is able to classify a given image region as face or non-face, estimate the head pose, locate face landmarks and recognize gender.}
	\label{fig:hf-resnet-model}
\end{figure*}

{\bf{R-CNN\_Pose:}} This model is used for head pose estimation task. The outputs of the network are roll, pitch and yaw of the face. Figure~\ref{fig:rcnns}(c) presents the network architecture.  Similar to R-CNN\_Fiducial, only region proposals with overlap$>0.5$ with the ground truth bounding box are used for training. The training loss is computed using \eqref{eq:pose_loss}.

{\bf{R-CNN\_Gender:}} This model is used for face gender recognition task. The network architecture is shown in Figure~\ref{fig:rcnns}(d). It has the same training set as R-CNN\_Fiducial and R-CNN\_Pose. The training loss is computed using \eqref{eq:gender_loss}.

{\bf{Multitask\_Face:}} Similar to HyperFace, this model is used to simultaneously detect face, localize landmarks, estimate pose and predict its gender. The only difference between Multitask\_Face and HyperFace is that HyperFace fuses the intermediate layers of the network whereas Multitask\_Face combines the tasks using the common fully connected layer at the end of the network as shown in Figure~\ref{fig:multitask_face}. Since it provides the landmarks and face score, it leverages iterative region proposals and landmark-based NMS post-processing algorithms during evaluation.

The performance of all the above models for their respective tasks are evaluated and discussed in details in Section~\ref{sec:res}.

\subsection{HyperFace-ResNet}

The CNN architectures have improved a lot over the years, mainly due to an increase in number of layers~\cite{he2016deep}, effective convolution kernel size~\cite{simonyan2014very}, batch normalization~\cite{ioffe2015batch} and skip connections. Recently, He et al.~\cite{he2016deep} proposed a deep residual network architecture with more than $100$ layers, that achieves state-of-the-art results on the ImageNet challenge~\cite{deng2009imagenet}. Hence, we propose a variant of HyperFace that is built using the ResNet-101~\cite{he2016deep} model instead of AlexNet~\cite{NIPS2012_4824}. The proposed network called HyperFace-ResNet (HF-ResNet) significantly improves upon its AlexNet baseline for all the tasks of face detection, landmarks localization, pose estimation and gender recognition. Figure~\ref{fig:hf-resnet-model} shows the network architecture for HF-ResNet.

Similar to HyperFace, we fuse the geometrically rich features from the lower layers and semantically strong features from the deeper layers of ResNet, such that multi-task learning can leverage from their synergy. Taking inspiration from~\cite{hu2016finding}, we fuse the features using hierarchical element-wise addition. Starting with `res2c' features, we first reduce its resolution using a $3 \times 3$ convolution kernel with stride of $2$. It is then passed through the a $1 \times 1$ convolution layer that increases the number of channels to match the next level features~ (`res3b3' in this case). Element-wise addition is applied between the two to generate a new set of fused features. The same operation is applied in a cascaded manner to fuse `res4b22' and `res5c' features of the ResNet-101 model. Finally, average pooling is carried out to generate $2048$-dimensional feature vector that is shared among all the tasks. Task-specific sub-networks are branched out separately in a similar way as HyperFace. Each convolution layer is followed by a Batch-Norm+Scale~\cite{ioffe2015batch} layer and ReLU activation unit. We do not use dropout in HF-ResNet. The training loss functions are the same as described in Section~\ref{training}. 

HF-ResNet is slower than HyperFace since it performs more convolutions. This makes it difficult to be used with Selective Search~\cite{vandeSande:2011:SSS:2355573.2356474} algorithm which generates more than $2000$ region proposals to be processed. Hence, we use a faster version of region proposals using high recall SSD~\cite{liu2016ssd} face detector. It produces $200$ proposals, needing just $0.05$s. This considerably reduces the total runtime for HF-ResNet to less than $1$s. The fast version of HyperFace is discussed in Section~\ref{sec:fast-hyperface}.

\section{Experimental Results}
\label{sec:res}
We evaluated the proposed HyperFace method, along with HF-ResNet, Multask\_Face, R-CNN\_Face, R-CNN\_Fiducial, R-CNN\_Pose and R-CNN\_Gender on six challenging datasets:
\begin{itemize}
\item
Annotated Face in-the-Wild (AFW)~\cite{AFW_dataset_CVPR2012} for evaluating face detection, landmarks localization, and pose estimation tasks
\item
300-W Faces in-the-wild~(IBUG)~\cite{sagonas2013300} for evaluating $68$-point landmarks localization.
\item
Annotated Facial Landmarks in the Wild (AFLW) \cite{AFLW} for evaluating landmarks localization and pose estimation tasks
\item
Face Detection Dataset and Benchmark (FDDB) \cite{fddbTech} and PASCAL faces \cite{PASCAL_faces} for evaluating the face detection results
\item
 Large-scale CelebFaces Attributes (CelebA) \cite{CelebA} and LFWA \cite{LFWTech} for evaluating gender recognition results.
 \end{itemize}
Our method was trained on randomly selected $20,997$ images from the AFLW dataset using Caffe \cite{jia2014caffe}. The remaining $1000$ images were used for testing.

\begin{figure*}[htp!]
	\centering
	\includegraphics[width=9.0cm, height=5.5cm]{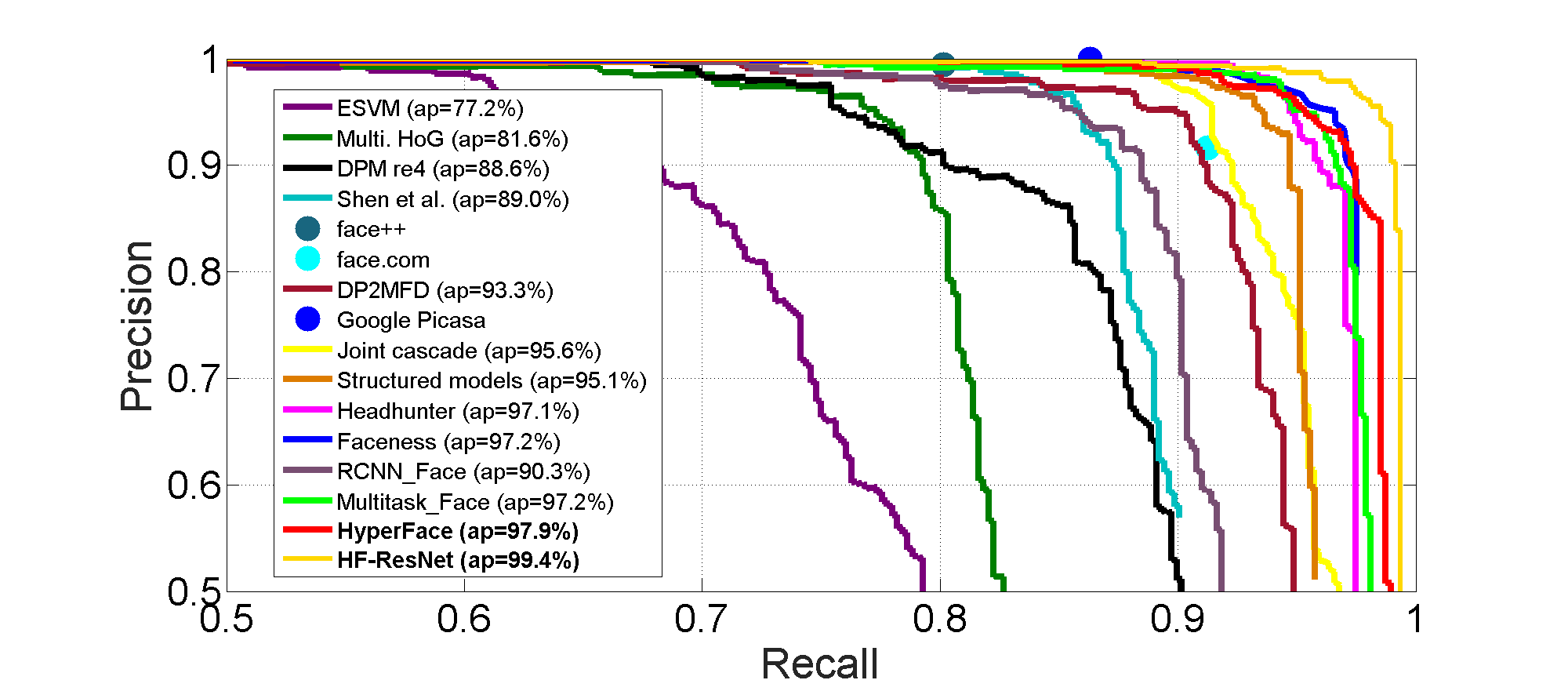}\hskip40pt\includegraphics[width=7.0cm, height=5.0cm]{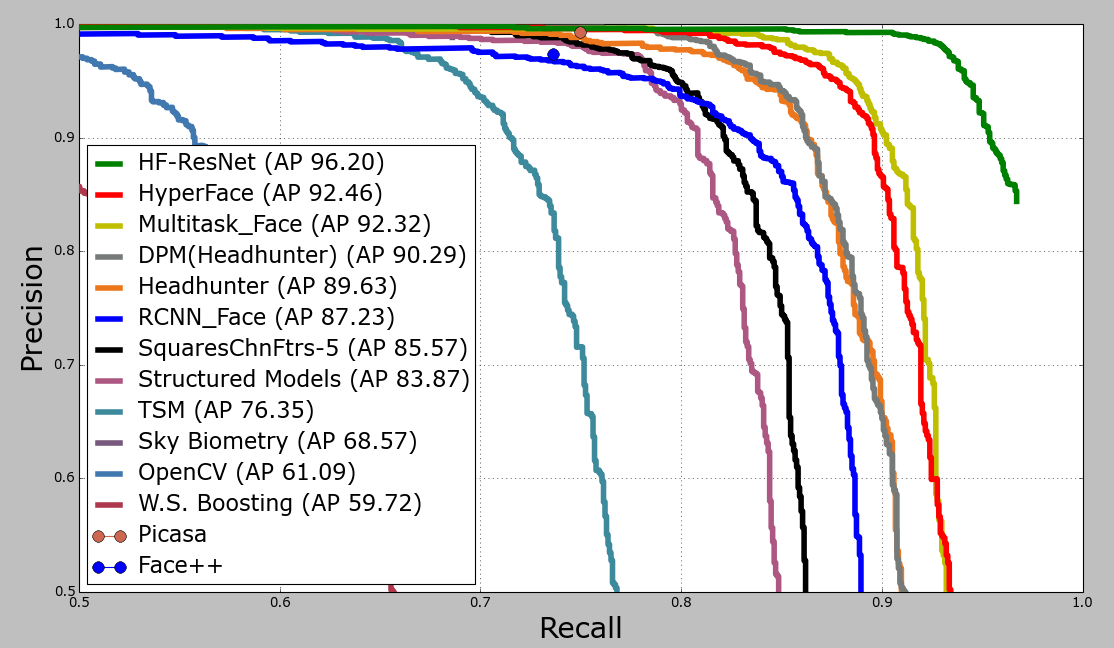}\\
	\hskip40pt(a)\hskip250pt(b)\\
	\caption{\textbf{Face Detection} performance evaluation on (a) the AFW dataset, (b) the PASCAL faces dataset. The numbers in the legend are the mean average precision (mAP) for the corresponding datasets.}
	\label{fig:detections}
\end{figure*} 

\subsection{Face Detection}
We present face detection results for AFW, PASCAL and FDDB datasets.  The AFW dataset~\cite{AFW_dataset_CVPR2012} was collected from Flickr and the images in this dataset contain large variations in appearance and viewpoint.  In total there are  205 images with 468 faces in this dataset.  The FDDB dataset~\cite{fddbTech} consists of 2,845 images containing 5,171 faces collected from news articles on the Yahoo website.  This dataset is the most widely used benchmark for unconstrained face detection.  The PASCAL faces dataset~\cite{PASCAL_faces} was collected from the test set of PASCAL person layout dataset, which is a subset from PASCAL VOC~\cite{PASCALVOC}.  This dataset contains 1335 faces from 851 images with large appearance variations. For improved face detection performance, we learn a SVM classifier on top of $fc_{detection}$ features using the training splits from the FDDB dataset. 

Some of the recent published methods compared in our evaluations include DP2MFD~\cite{FD_BTAS2015}, Faceness~\cite{faceness_ICCV2015}, HeadHunter~\cite{HeadHunter_Mathias_ECCV2014}, JointCascade~\cite{JointCascade_LI_ECCV2014}, CCF~\cite{CCF_ICCV2015}, SquaresChnFtrs-5~\cite{HeadHunter_Mathias_ECCV2014}, CascadeCNN~\cite{CascadeCNN_CVPR2015}, Structured Models~\cite{PASCAL_faces}, DDFD~\cite{DDFD_ICMR2015}, NPDFace~\cite{NPDFace_PAMI2015}, PEP-Adapt~\cite{PEP_Adapt_LI_ICCV2013}, TSM~\cite{AFW_dataset_CVPR2012}, as well as three commercial systems Face++, Picasa and Face.com. 

\begin{figure}[htp!]
\centering
\includegraphics[width=9.0cm, height=5.8cm]{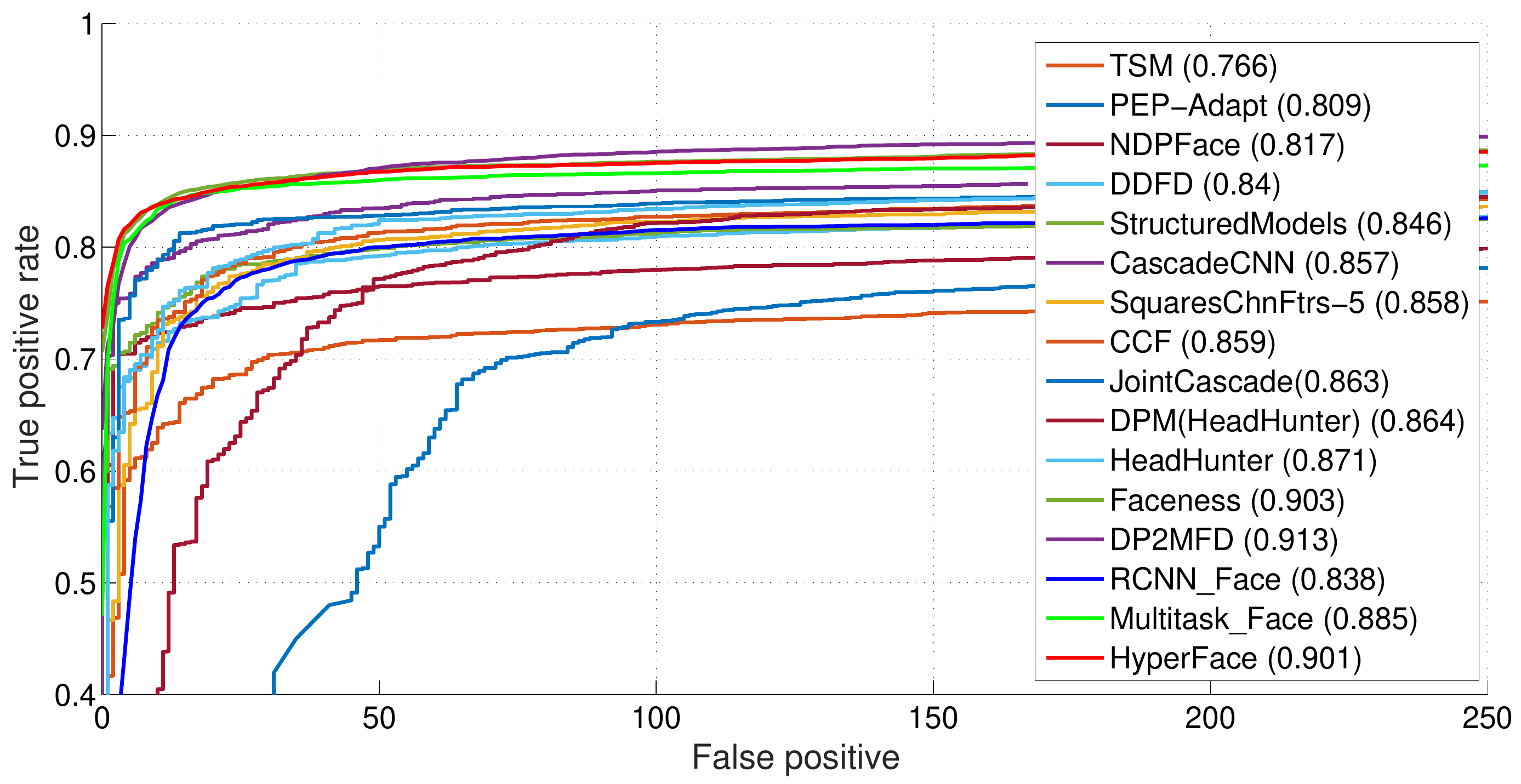}
\caption{\textbf{Face Detection} performance evaluation on the FDDB dataset. The numbers in the legend are the mean average precision.}
\label{fig:fddb}
\end{figure}

The precision-recall curves of different detectors corresponding to AFW and PASCAL faces datasets are shown in Figures~\ref{fig:detections} (a) and (b), respectively.    Figure~\ref{fig:fddb} compares the performance of different detectors using the Receiver Operating Characteristic (ROC) curves on the FDDB dataset.  As can be seen from these figures, both HyperFace and HF-ResNet outperform all the reported academic and commercial detectors on the AFW and PASCAL datasets. HyperFace achieves a high mean average precision ($mAP$) of $97.9\%$ and $92.46\%$, for AFW and PASCAL datasets respectively. HF-ResNet further improves the mAP to $99.4\%$ and $96.2\%$ respectively.

 The FDDB dataset is very challenging for HyperFace and any other R-CNN-based face detection methods, as the dataset contains many small and blurred faces. First, some of these faces do not get included in the region proposals from selective search. Second, re-sizing small faces to the input size of $227 \times 227$ adds distortion to the face resulting in  low detection score. In spite of these issues, HyperFace performance is comparable to recently published deep learning-based face detection methods such as DP2MFD~\cite{FD_BTAS2015} and Faceness~\cite{faceness_ICCV2015}  on the FDDB dataset~\footnote{http://vis-www.cs.umass.edu/fddb/results.html} with $mAP$ of $90.1\%$.

It is interesting to note the performance differences between R-CNN\_Face, Multitask\_Face and HyperFace for the face detection tasks. Figures~\ref{fig:detections}, and \ref{fig:fddb} clearly show that multitask CNNs (Multitask\_Face and HyperFace) outperform R-CNN\_Face by a wide margin. The boost in the performance gain is mainly due to the following two reasons. First, multitask learning approach helps the network to learn improved features for face detection which is evident from their $mAP$ values on the AFW dataset. Using just the linear bounding box regression and traditional NMS, the HyperFace obtains a $mAP$ of $94\%$ (Figure~\ref{fig:post_processing}) while R-CNN\_Face achieves a $mAP$ of $90.3\%$. Second, having landmark information associated with detection boxes makes it easier to localize the bounding box to a face, by using IRP and L-NMS algorithms. On the other hand, HyperFace and Multi-task\_Face perform comparable to each other for all the face detection datasets which suggests that the network does not gain much by fusing intermediate layers for the face detection task.

%the reason is that fddb has many small and blurred faces... but our algorithm has a bit of limitation to it that it needs to scale up every box to 227x227 thats why it is difficult for very smalll images

 \begin{figure}[htp!]
 \centering
\includegraphics[width=9.0cm]{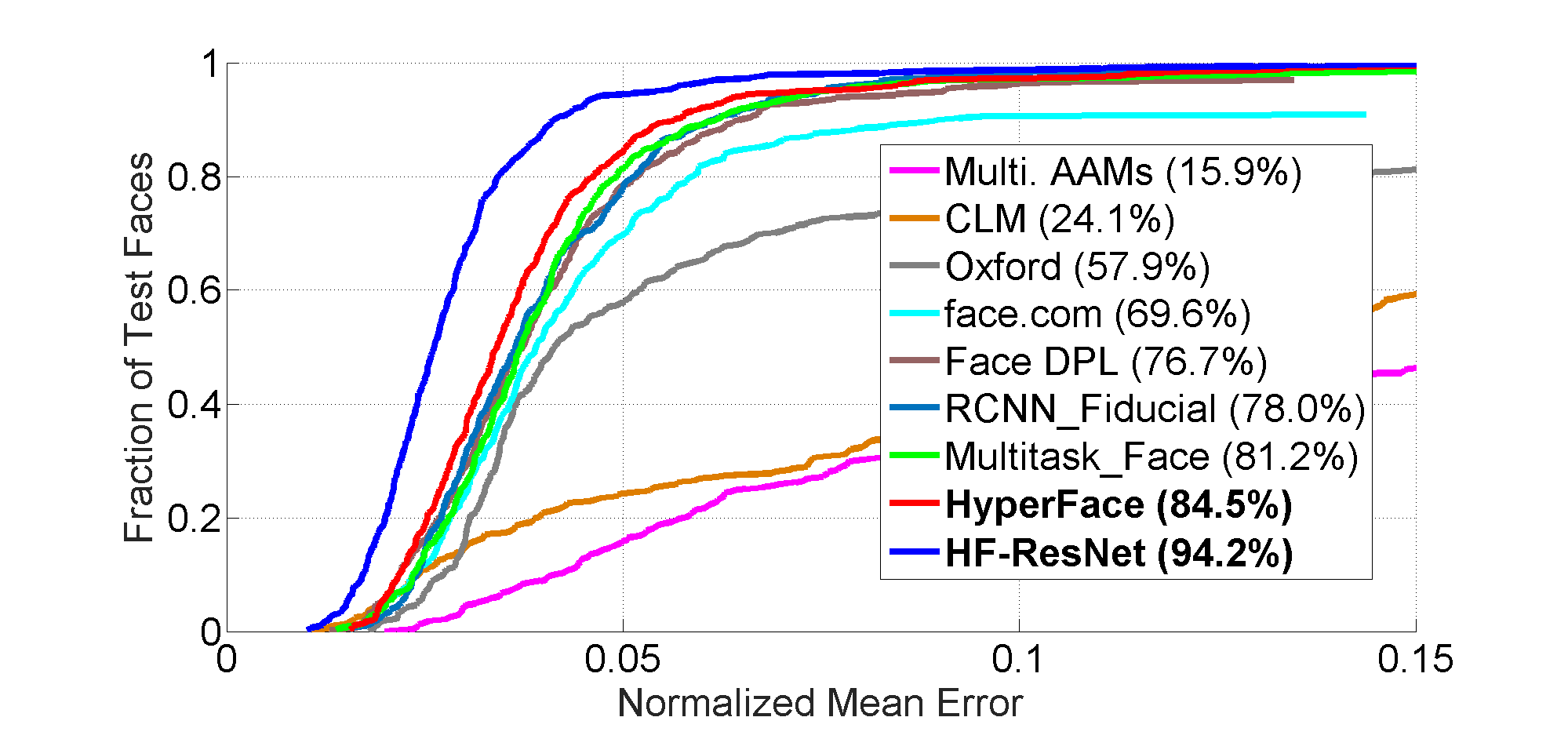}
\caption{\textbf{Landmarks Localization} cumulative error distribution curves on the AFW dataset. The numbers in the legend are the fraction of testing faces that have average error below (5\%) of the face size.}
\label{fig:landmark_afw}
\end{figure}

\subsection{Landmarks Localization}
\label{res:landmarks}
We evaluate the performance of different landmarks localization algorithms on AFW~\cite{AFW_dataset_CVPR2012} and AFLW~\cite{AFLW} datasets. Both of these datasets contain faces with full pose variations. Some of the methods compared include Multiview Active Appearance Model-based method (Multi. AAM)~\cite{AFW_dataset_CVPR2012}, Constrained Local Model (CLM)~\cite{CLM}, Oxford facial landmark detector \cite{Buffy}, Zhu~\cite{AFW_dataset_CVPR2012}, FaceDPL~\cite{FaceDPL}, JointCascade~\cite{JointCascade_LI_ECCV2014}, CDM~\cite{yu2013pose}, RCPR~\cite{burgos2013robust}, ESR~\cite{DBLP:journals/ijcv/CaoWWS14}, SDM~\cite{XiongD13} and 3DDFA~\cite{DBLP:journals/corr/ZhuLLSL15}. Although both of these datasets provide ground truth bounding boxes, we do not use them for evaluating on HyperFace, HF-ResNet, Multitask\_Face and R-CNN\_Fiducial. Instead we use the respective algorithms to detect both the face and its fiducial points. Since, the R-CNN\_Fiducial cannot detect faces, we provide it with the detections from the HyperFace.

 Figure~\ref{fig:landmark_afw} compares the performance of different landmark localization methods on the AFW dataset using the protocol defined in~\cite{FaceDPL}. In this figure, (*) indicates that models that are evaluated on near frontal faces  or use hand-initialization~\cite{AFW_dataset_CVPR2012}. The dataset provides six keypoints for each face which are: left\_eye\_center, right\_eye\_center, nose\_tip, mouth\_left, mouth\_center and mouth\_right.  We compute the error as the mean distance between the predicted and ground truth keypoints, normalized by the face size. The plots for comparison were obtained from~\cite{FaceDPL}.

 \begin{figure}[htp!]
 \centering
\includegraphics[width=9.0cm, height=6.0cm]{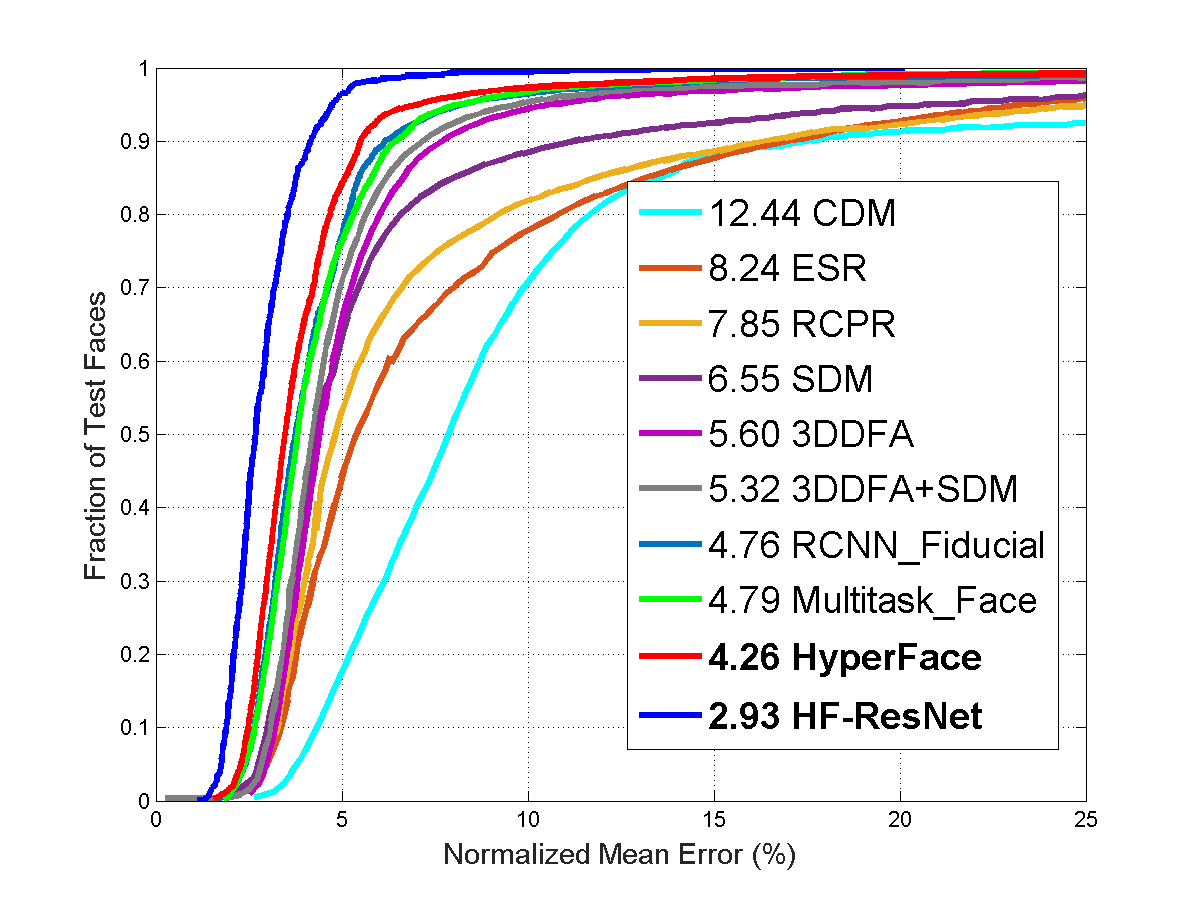}
\caption{\textbf{Landmarks Localization} cumulative error distribution curves on the AFLW dataset. The numbers in the legend are the average NME for the test images. The test samples are chosen such that samples with absolute yaw angles between [$0^{\circ}$,$30^{\circ}$], [$30^{\circ}$,$60^{\circ}$] and [$60^{\circ}$,$90^{\circ}$] are $1/3$ each.}
\label{fig:landmark_aflw}
\end{figure}

For the AFLW dataset, we calculate the error using all the visible keypoints. For AFW, we adopt the same protocol as defined in \cite{DBLP:journals/corr/ZhuLLSL15}. The only difference is that our AFLW testset consists of only $1000$ images with $1132$ face samples, since we use the rest of the images for training. To be consistent with the protocol, we randomly create a subset of $450$ samples from our testset whose absolute yaw angles within [$0^{\circ}, 30^{\circ}$], [$30^{\circ}, 60^{\circ}$] and [$60^{\circ}, 90^{\circ}$] are $1/3$ each. Figure~\ref{fig:landmark_aflw} compares the performance of different landmark localization methods. We obtain the comparison plots from~\cite{DBLP:journals/corr/ZhuLLSL15} where the evaluations for RCPR, ESR and SDM are carried out after adapting the algorithms to face profiling. Table~\ref{tab:aflw} provides the Normalized Mean Error (NME) for AFLW dataset, for each of the pose group.

\begin{table}[htp!]
\centering
\caption{The NME(\%) of face alignment results on AFLW test set with the best results highlighted.}
\label{tab:aflw}
\begin{tabular}{|c|c|c|c|c|c|}
\hline
                   & \multicolumn{5}{c|}{AFLW Dataset (21 pts)}                                    \\ \hline
Method             & {[}0, 30{]}   & {[}30, 60{]}  & {[}60, 90{]}  & mean          & std           \\ \hline
CDM~\cite{yu2013pose}                & 8.15          & 13.02         & 16.17         & 12.44         & 4.04          \\ \hline
RCPR~\cite{burgos2013robust}         & 5.43          & 6.58          & 11.53         & 7.85          & 3.24          \\ \hline
ESR~\cite{DBLP:journals/ijcv/CaoWWS14}                & 5.66          & 7.12          & 11.94         & 8.24          & 3.29          \\ \hline
SDM~\cite{XiongD13}                & 4.75          & 5.55          & 9.34          & 6.55          & 2.45          \\ \hline
3DDFA~\cite{DBLP:journals/corr/ZhuLLSL15}              & 5.00          & 5.06          & 6.74          & 5.60          & 0.99          \\ \hline
3DDFA~\cite{DBLP:journals/corr/ZhuLLSL15}+SDM          & 4.75          & 4.83          & 6.38          & 5.32          & 0.92          \\ \hline
R-CNN\_Fiducial     & 4.49          & 4.70          & 5.09          & 4.76          & 0.30 \\ \hline
Multitask\_Face    & 4.20          & 4.93          & 5.23          & 4.79          & 0.53          \\ \hline \hline
\textbf{HyperFace} & \textbf{3.93} & \textbf{4.14} & \textbf{4.71} & \textbf{4.26} & \textbf{0.41}          \\ \hline
\textbf{HF-ResNet} & \textbf{2.71} & \textbf{2.88} & \textbf{3.19} & \textbf{2.93} & \textbf{0.25}          \\ \hline
\end{tabular}
\end{table}

\begin{figure*}[htp!]
 \centering
\includegraphics[width=5.5cm, height=3.5cm]{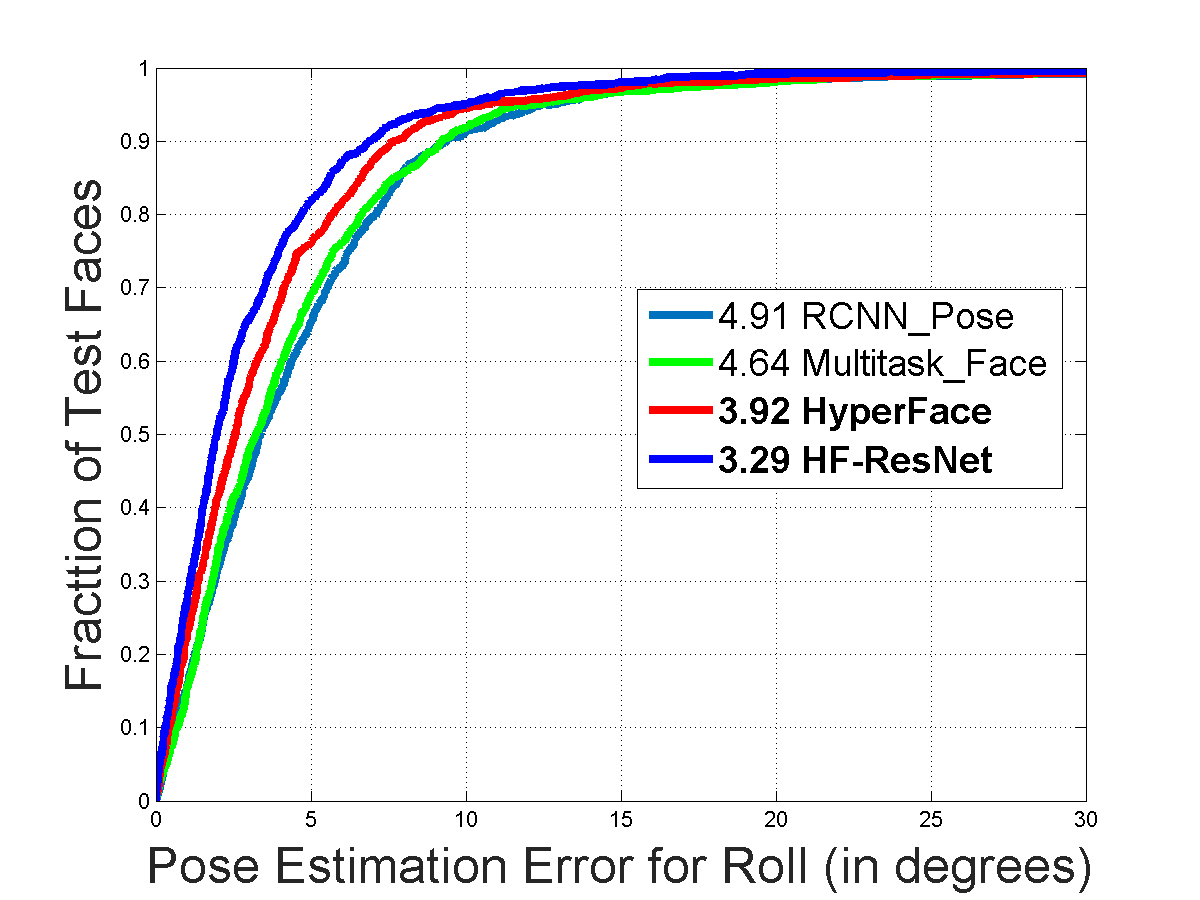}\hskip10pt\includegraphics[width=5.5cm, height=3.5cm]{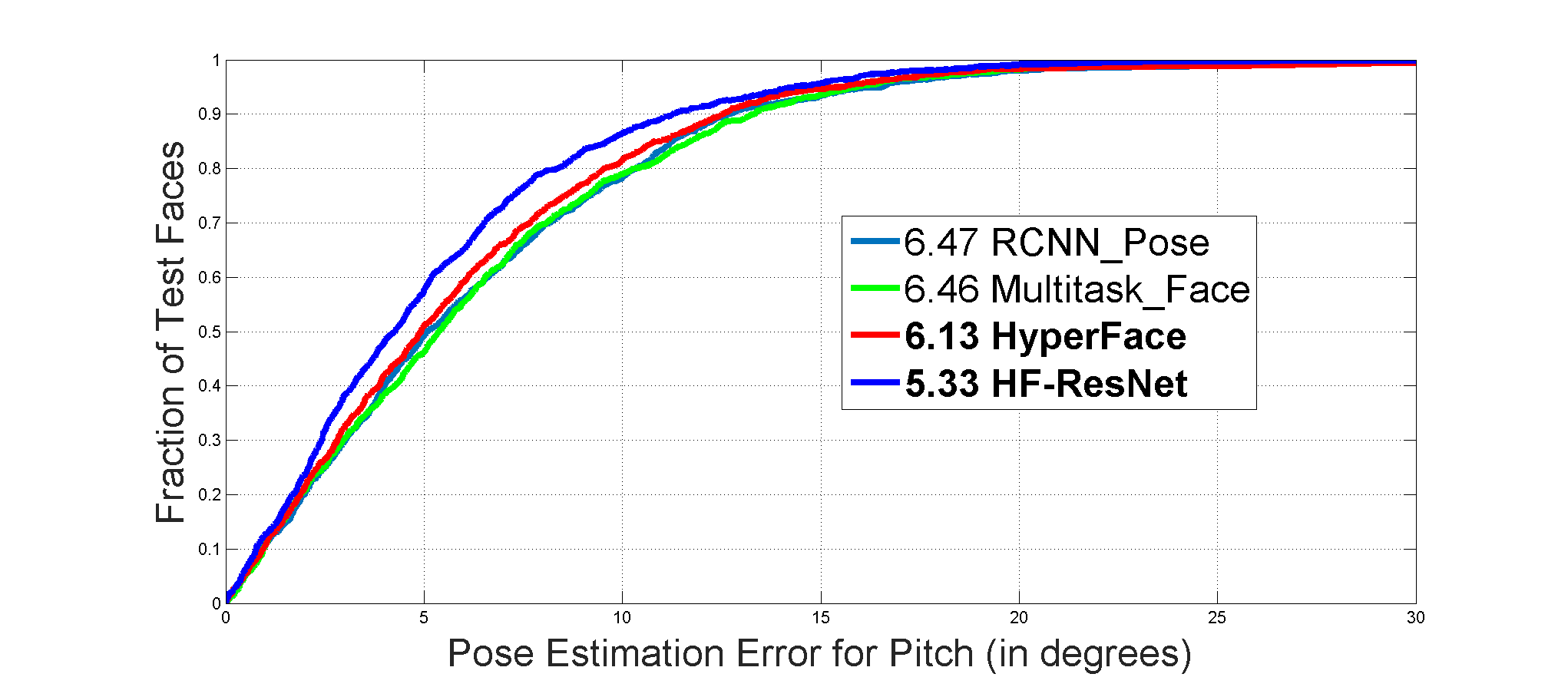}\hskip10pt\includegraphics[width=5.5cm, height=3.5cm]{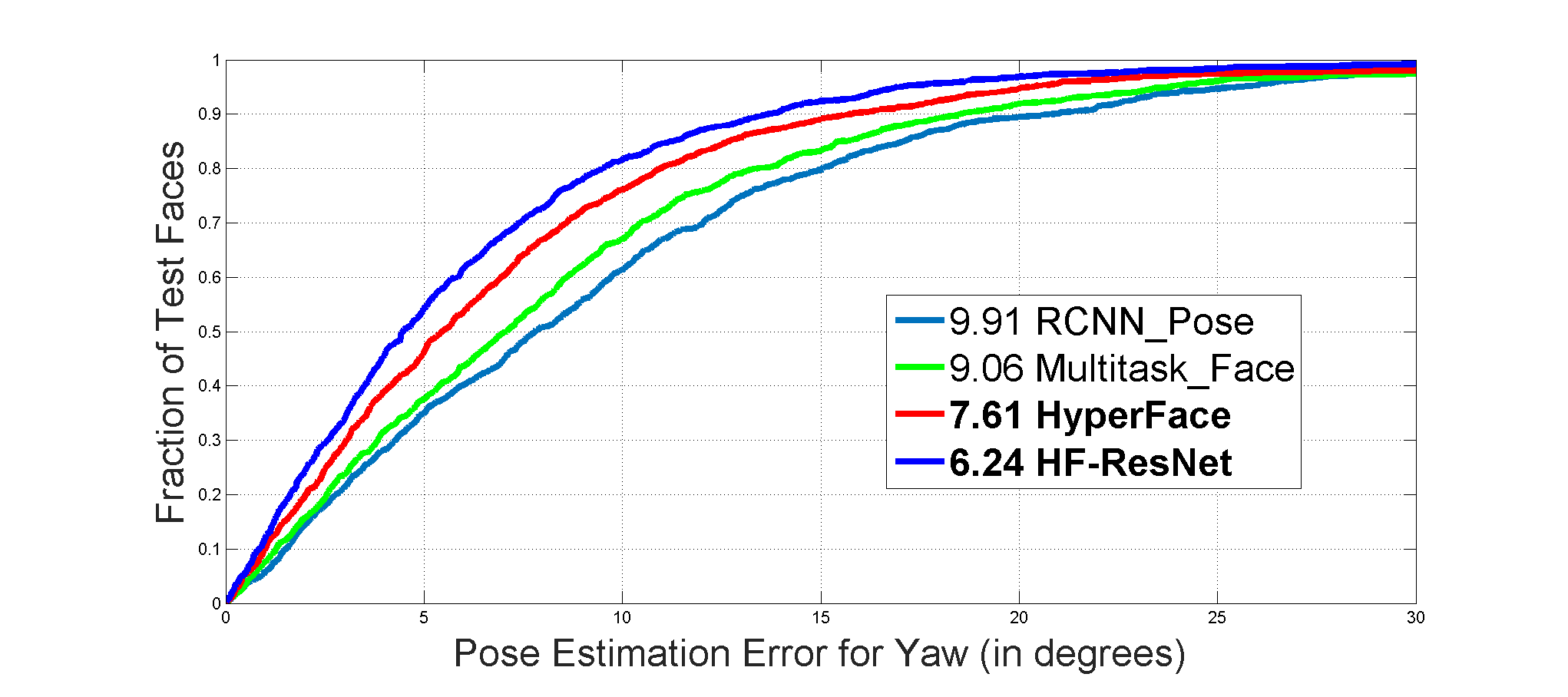}\\
(a)\hskip160pt(b)\hskip160pt(c)\\
\caption{\textbf{Pose Estimation} performance evaluation on AFLW dataset for (a) roll (b) pitch and (c) yaw angles. The numbers in the legend are the mean error in degrees for the respective pose angles.}
\label{fig:pose_AFLW}
\end{figure*}

As can be seen from the figures, R-CNN\_Fiducial, Multitask\_Face, HyperFace and HF-ResNet outperform many recent state-of-the-art landmark localization methods including FaceDPL~\cite{FaceDPL}, 3DDFA~\cite{DBLP:journals/corr/ZhuLLSL15} and SDM~\cite{XiongD13}. Table~\ref{tab:aflw} shows that HyperFace performs consistently accurate over all pose angles. This clearly suggests that while most of the methods work well on frontal faces, HyperFace is able to predict landmarks for faces with full pose variations. Moreover, we find that R-CNN\_Fiducial and Multitask\_Face attain similar performance. The HyperFace has an advantage over them as it uses the intermediate layers for fusion. The local information is contained well in the lower layers of CNN and becomes invariant as depth increases. Fusing the layers brings out that hidden information which boosts the performance for the landmark localization task. Additionally, we observe that HF-ResNet significantly improves the performance over HyperFace for both AFW and AFLW datasets. The large margin in performance can be attributed to the larger depth for the HF-ResNet model.

We also evaluate our models on the challenging subset of the 300-W~\cite{sagonas2013300} landmarks localization dataset~(IBUG). The dataset contains $135$ test images with wide variations in expression and illumination. The head-pose angle varies from $-60^{\circ}$ to $60^{\circ}$ in yaw. Since the dataset contains $68$ landmarks points instead of $21$ used in AFLW~\cite{AFLW} training, the model cannot be directly applied for evaluating IBUG. We retrain the network for predicting $68$ facial key-points as a separate task in conjunction with the proposed tasks in hand. We implement it by adding two fully-connected layers in a cascade manner to the shared feature space (fc-full), having dimensions $512$ and $136$, respectively. 

Following the protocol described in~\cite{ren2014face}, we use $3,148$ faces with $68$-point annotations for training. The network is trained end-to-end for the localization of $68$-points landmarks along with the other tasks mentioned in Section~\ref{training}. We use the standard Euclidean loss function for training. For evaluation, we compute the average error of all $68$ landmarks normalized by the inter-pupil distance. Table~\ref{tbl:ibug} compares the Normalized Mean Error (NME) obtained by HyperFace and HF-ResNet with other recently published methods. We observe that HyperFace achieves a comparable NME of $10.88$, while HF-ResNet achieves the state-of-the-art result on IBUG~\cite{sagonas2013300} with NME of $8.18$. This shows the effectiveness of the proposed models for $68$-point landmarks localization. 

\begin{table}[htp!]
	\begin{center}
		\caption{Normalized Mean Error (in \%) of $68$-point landmarks localization on IBUG~\cite{sagonas2013300} dataset. }
		\label{tbl:ibug}
		
				\begin{tabular}{|l|c|}
					\hline
					Method & Normalized Mean Error \\
					\hline\hline
					CDM~\cite{yu2013pose}&19.54\\ \hline
					RCPR~\cite{burgos2013robust}&17.26\\ \hline
					ESR~\cite{DBLP:journals/ijcv/CaoWWS14}&17.00\\ \hline
					SDM~\cite{XiongD13}&15.40\\ \hline
					LBF~\cite{ren2014face}&11.98\\ \hline
					LDDR~\cite{DBLP:journals/corr/KumarRPC16}&11.49\\ \hline
					CFSS~\cite{zhu2015face}&9.98\\ \hline
					3DDFA~\cite{DBLP:journals/corr/ZhuLLSL15}&10.59\\ \hline
					TCDCN~\cite{zhang2016learning}&8.60\\
					\hline
					\hline
					HyperFace&10.88\\ \hline
					HF-ResNet&\bf{8.18}\\
					\hline
				\end{tabular}
		
	\end{center}
\end{table}

\subsection{Pose Estimation}
\begin{figure}[htp!]
 \centering
\includegraphics[width=8.5cm, height=5.5cm]{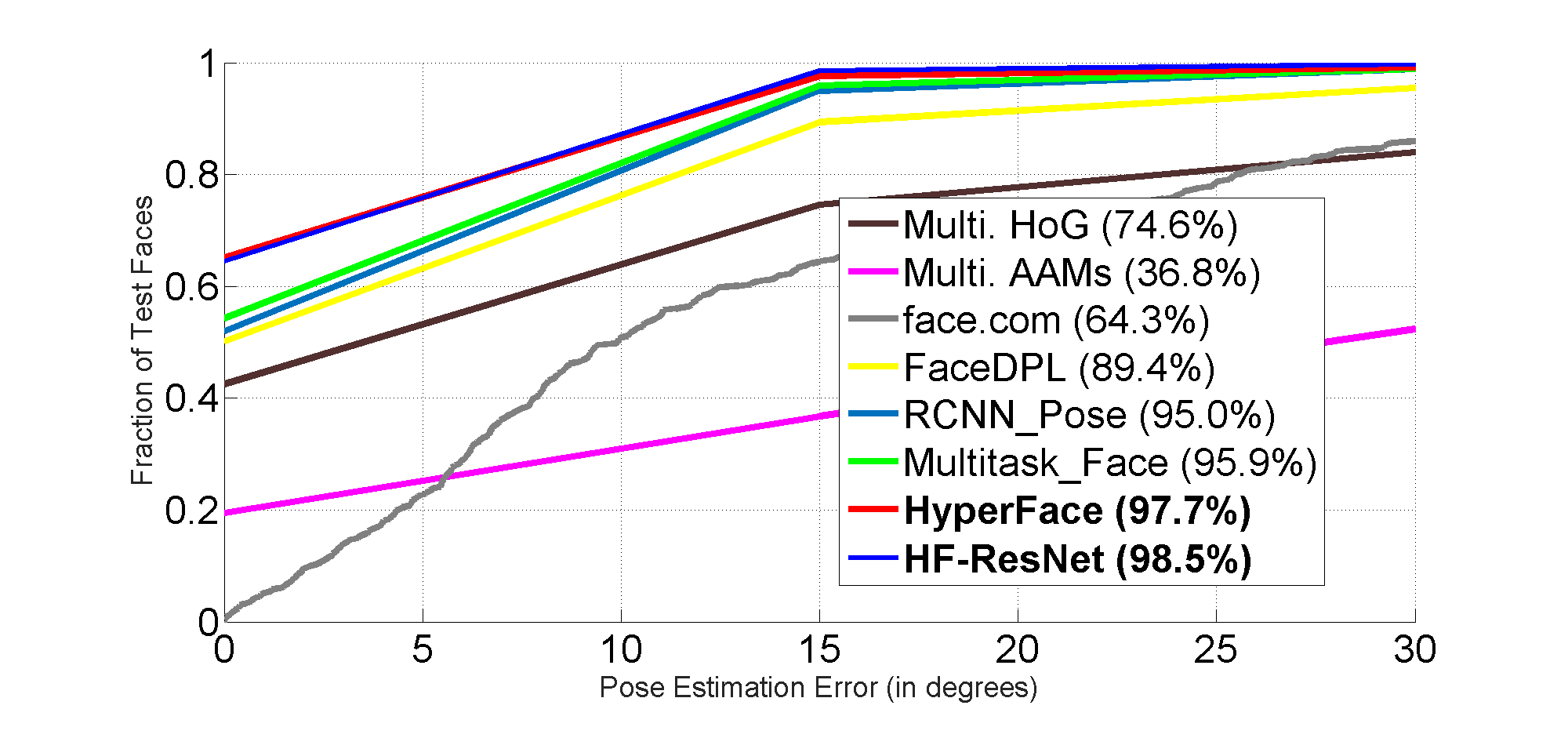}\\
\caption{\textbf{Pose Estimation} cumulative error distribution curves on AFW dataset. The numbers in the legend are the percentage of faces that are labeled within $\pm15^{\circ}$  error tolerance.}
\label{fig:pose_afw}
\end{figure}

We evaluate R-CNN\_Pose, Multitask\_Face and HyperFace  on the AFW~\cite{AFW_dataset_CVPR2012} and AFLW~\cite{AFLW} datasets for the pose estimation task. The detection boxes used for evaluating the landmark localization task are used here as well for initialization. For the AFW dataset, we compare our approach with Multi. AAM~\cite{AFW_dataset_CVPR2012}, Multiview HoG \cite{AFW_dataset_CVPR2012}, FaceDPL\footnote{Available at: \url{http://www.ics.uci.edu/~dramanan/software/face/face_journal.pdf}} \cite{FaceDPL} and face.com.  Note that multiview AAMs are initialized using the ground truth bounding boxes (denoted by *).     Figure~\ref{fig:pose_afw} shows the cumulative error distribution curves on AFW dataset.    The curve provides the fraction of faces for which the estimated pose is within some error tolerance.  As can be seen from the figure, both HyperFace and HF-ResNet outperform existing methods by a large margin. For the AFLW dataset, we do not have pose estimation evaluation for any previous method. Hence, we show the performance of our method for different pose angles: roll, pitch and yaw in Figure~\ref{fig:pose_AFLW}~(a), (b) and (c) respectively. It can be seen that the network is able to learn roll, and pitch information better than yaw.

The performance traits of R-CNN\_Pose, Multitask\_Face, HyperFace and HF-ResNet for pose estimation task are similar to that of the landmarks localization task. R-CNN\_Pose and Multitask\_Face perform comparable to each other whereas HyperFace achieves a boosted performance due to the intermediate layers fusion. It shows that tasks which rely on the structure and orientation of the face work well with features from lower layers of the CNN. HF-ResNet further improves the performance for roll, pitch as well as yaw.

\subsection{Gender Recognition}
We present the gender recognition performance on CelebA~\cite{CelebA} and LFWA~\cite{LFWTech} datasets since these datasets come with gender information.  The CelebA and LFWA datasets contain labeled images selected from the CelebFaces \cite{CelebFaces_nips2014} and LFW \cite{LFWTech} datasets, respectively \cite{CelebA}.  The CelebA dataset contains 10,000 identities and there are 200,000 images in total.  The LFWA dataset has 13,233 images of 5,749 identities.  We compare our approach with FaceTracer \cite{facetracer}, PANDA-w \cite{PANDA}, PANDA-1 \cite{PANDA}, \cite{surfCascade_CVPR2013} with ANet and \cite{CelebA}.  The gender recognition performance of different methods is reported in Table~\ref{tbl:gender}.  On the LFWA dataset, our method outperforms PANDA \cite{PANDA} and FaceTracer \cite{facetracer}, and is equal to \cite{CelebA}.  On the CelebA dataset, our method performs comparably to \cite{CelebA}.  
Unlike \cite{CelebA} which uses $180,000$ images for training and validation, we only use $20,000$ images from validation set of CelebA to fine-tune the network. 

\begin{table}[htp!]
\begin{center}
\caption{Performance comparison (in \%) of \textbf{gender recognition} on CelebA and LFWA datasets. }\label{tbl:gender}
\resizebox{1.5\textwidth}{!}{\begin{minipage}{\textwidth}
\begin{tabular}{|l|c|c|}
\hline
Method & CelebA&LFWA \\
\hline\hline
FaceTracer \cite{facetracer}&91&84\\ \hline
PANDA-w \cite{PANDA}&93&86\\ \hline
PANDA-1 \cite{PANDA}&97&92\\ \hline
\cite{surfCascade_CVPR2013}+ANet & 95&91 \\ \hline
LNets+ANet \cite{CelebA} & \bf{98} &\bf{94}\\ \hline
R-CNN\_Gender &95 &91\\ \hline
Multitask\_Face &97 &93\\ \hline \hline
HyperFace&97  &\bf{94}\\ \hline
HF-ResNet&\bf{98}  &\bf{94}\\
\hline
\end{tabular}
\end{minipage}}
\end{center}
\end{table}

Similar to the face detection task, we find that gender recognition performs better for HyperFace and Multitask\_Face as compared to R-CNN\_Gender proving that learning related tasks together improves the discriminating capability of the individual tasks. Again, we do not see much difference in the performance of Multitask\_Face and HyperFace suggesting intermediate layers do not contribute much for the gender recognition task. HF-ResNet achieves state-of-the-art results on both CelebA~\cite{CelebA} and LFWA~\cite{LFWTech} datasets. 

\subsection{Effect of Post-Processing}
\begin{figure}[htp!]
 \centering
\includegraphics[width=8.5cm, height=5.5cm]{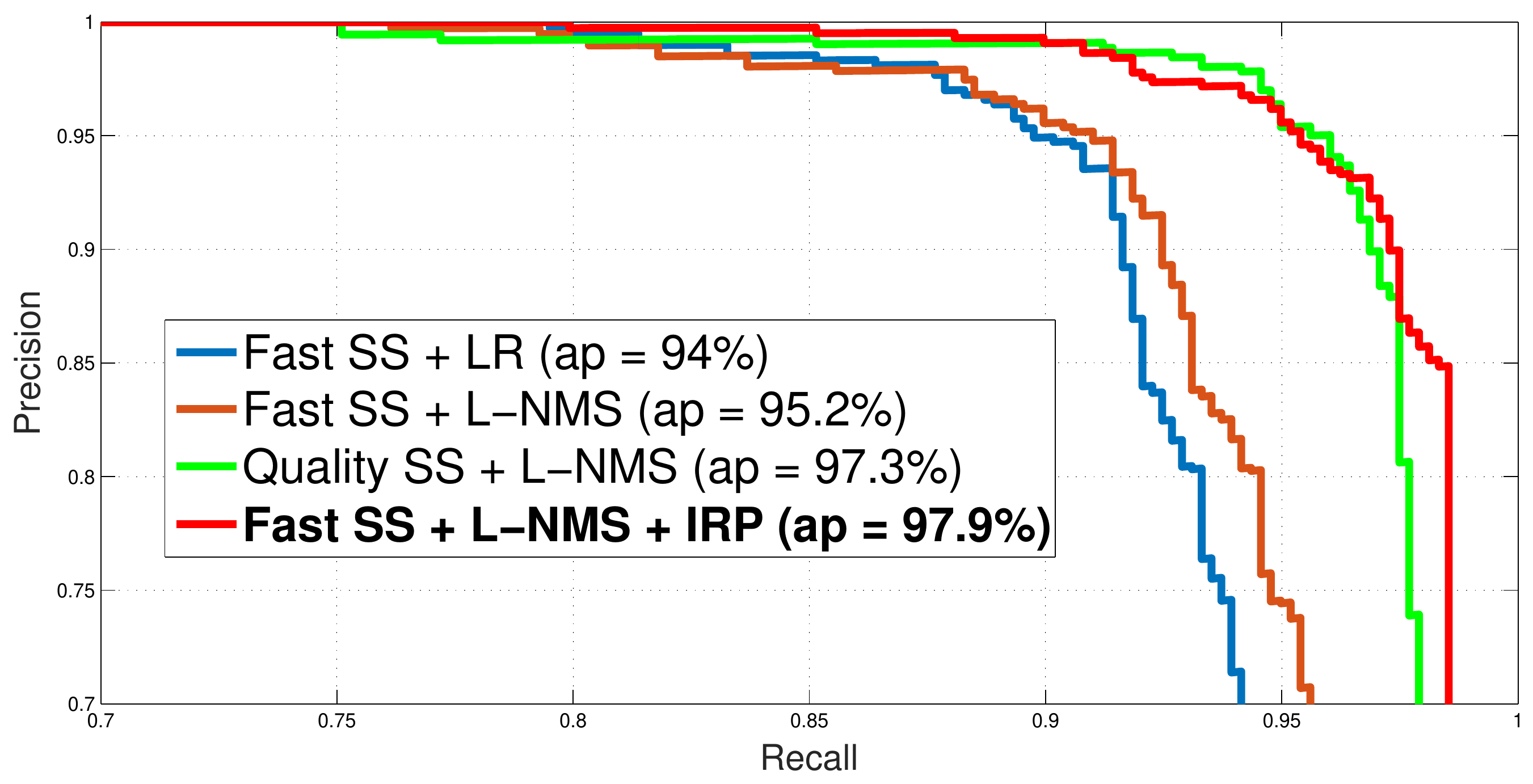}\\
\caption{Variations in performance of HyperFace with respect to the Iterative Region Proposals and Landmarks-based NMS. The numbers in the legend are the mean average precision.}
\label{fig:post_processing}
\end{figure}

Figure~\ref{fig:post_processing} provides an experimental analysis of the post-processing methods: IRP and L-NMS, for face detection task on the AFW dataset. \textit{Fast SS} denotes the quick version of selective search which produces around $2000$ region proposals and takes $2s$ per image to compute. On the other hand, \textit{Quality SS} refers to its slow version which outputs more than $10,000$ region proposals consuming more than $10s$ for one image. The HyperFace with a linear bounding box regression and traditional NMS achieves a $mAP$ of $94\%$. Just by replacing them with L-NMS provides a boost of $1.2\%$. In this case, bounding-box is constructed using the landmarks information rather linear regression. Additionaly, we can see from the figure that although \textit{Quality SS} generates more region proposals, it performs worse than \textit{Fast SS} with iterative region proposals. IRP adds $300$ new regions for a typical image consuming less than $0.5s$ which makes it highly efficient as compared to \textit{Quality SS}.  

\begin{figure*}[htp!]
 \centering
 \includegraphics[width=16cm]{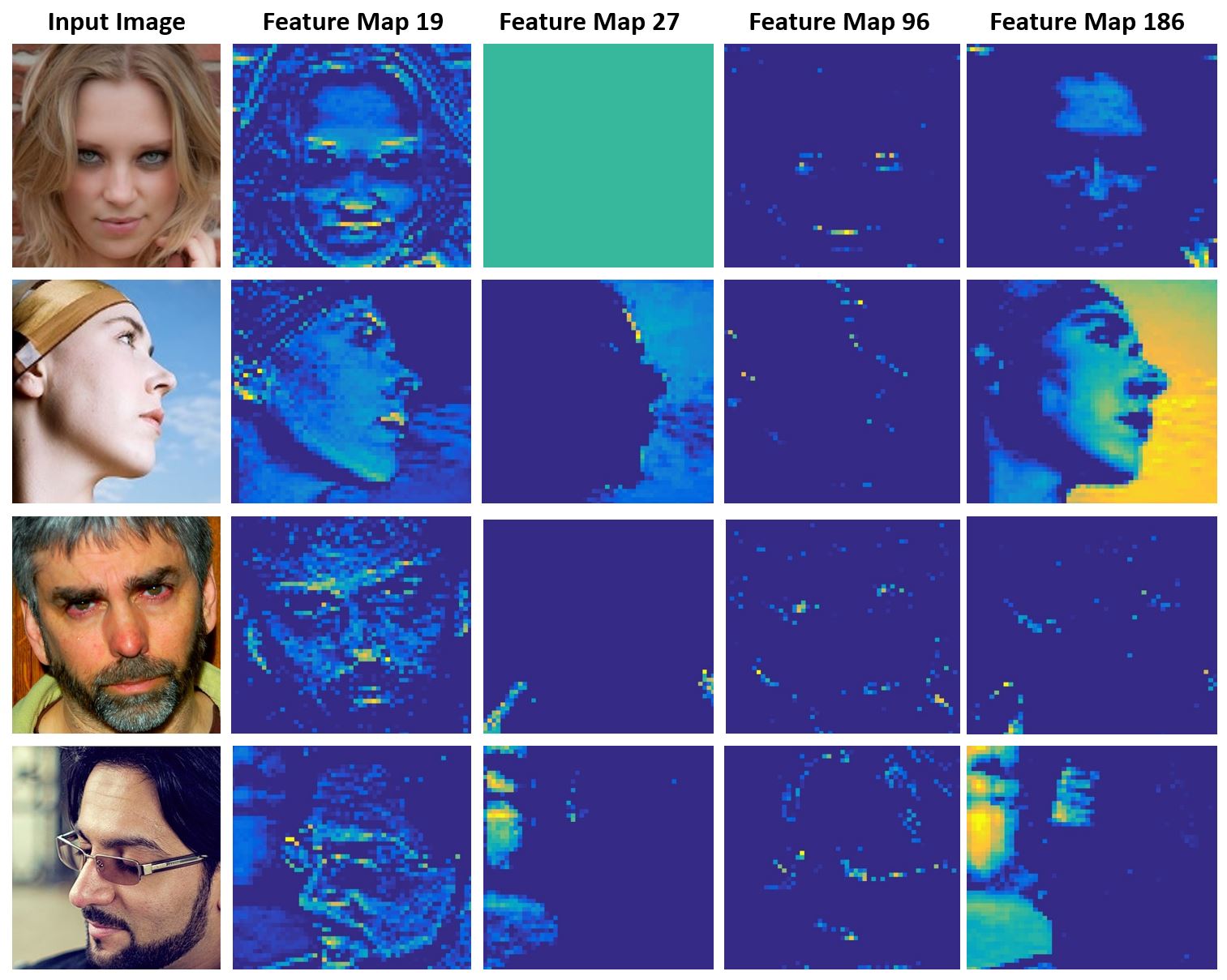}\\
 \caption{Activations of selected feature maps from \textbf{conv\_all}  layer of the HyperFace architecture. Green and yellow colors denote high activation whereas blue denotes low activation units. These features depict the distinguishable face traits for the tasks of face detection, landmarks localization, pose estimation and gender recognition.}
\label{fig:vizualization}
\end{figure*}

\begin{figure*}[htp!]
 \centering
\includegraphics[width=5.5cm, height=3.85cm]{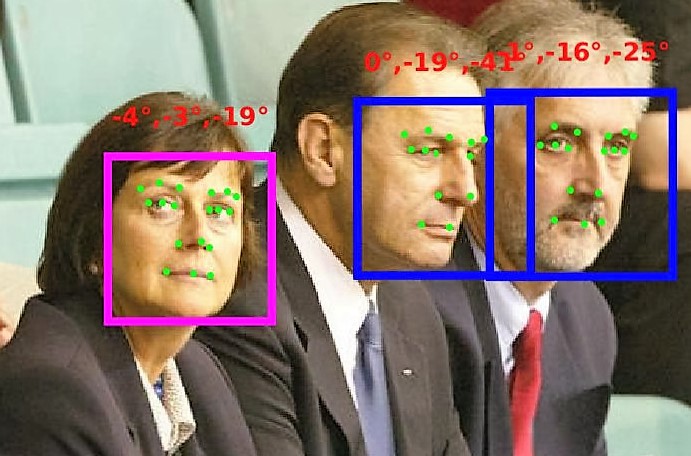}\hskip5pt\includegraphics[width=5.5cm,height=3.85cm]{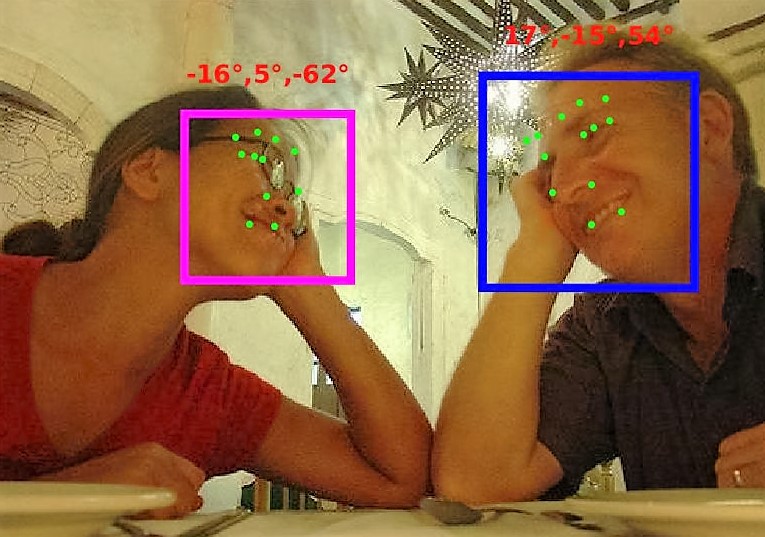}\hskip5pt\includegraphics[width=5.5cm,height=3.85cm]{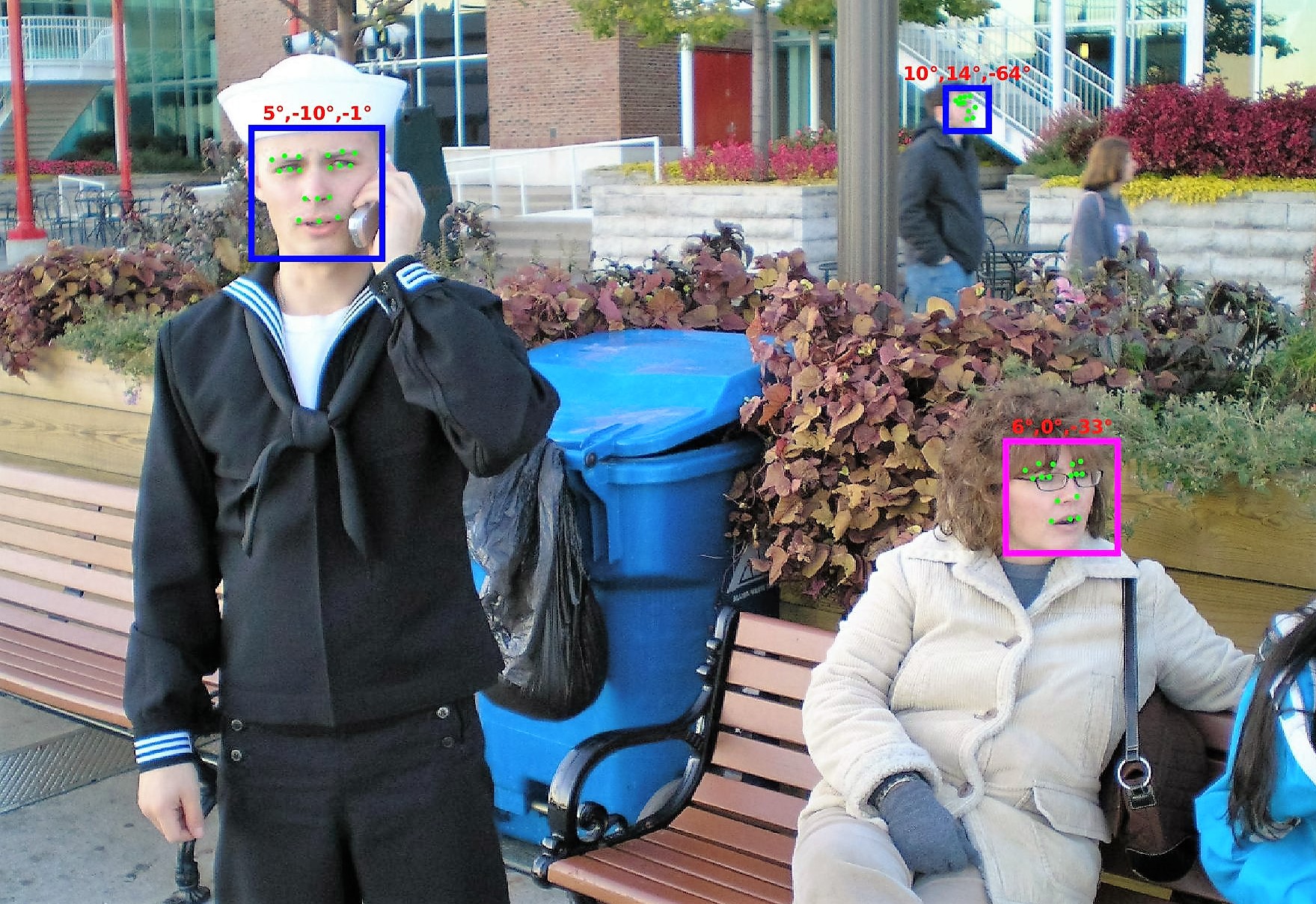}\\
\vskip5pt
\includegraphics[width=5.5cm, height=3.85cm]{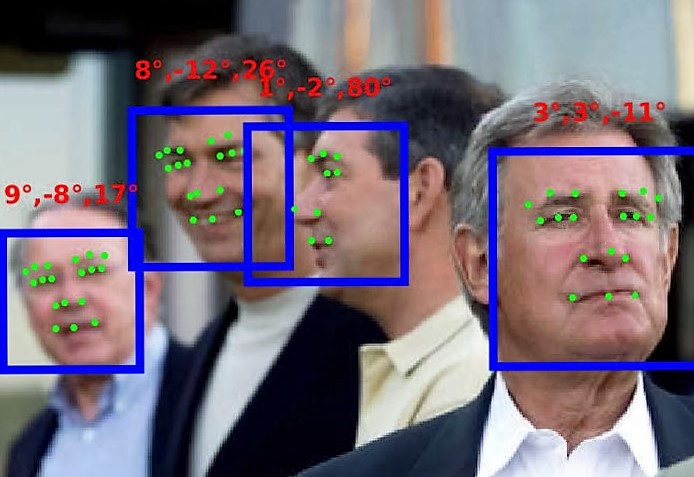}\hskip5pt\includegraphics[width=5.5cm,height=3.85cm]{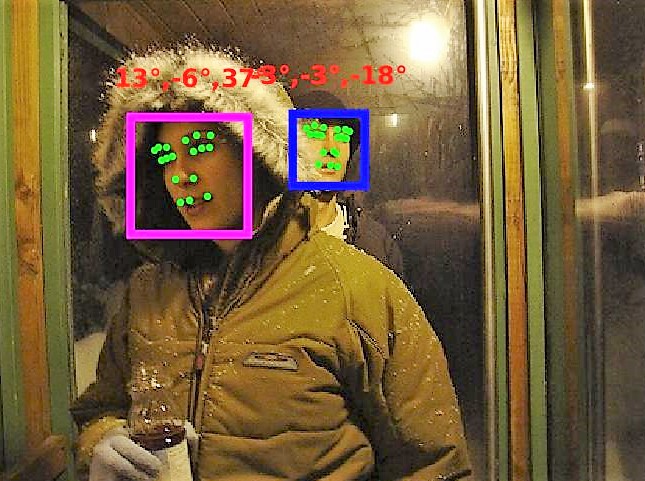}\hskip5pt\includegraphics[width=5.5cm,height=3.85cm]{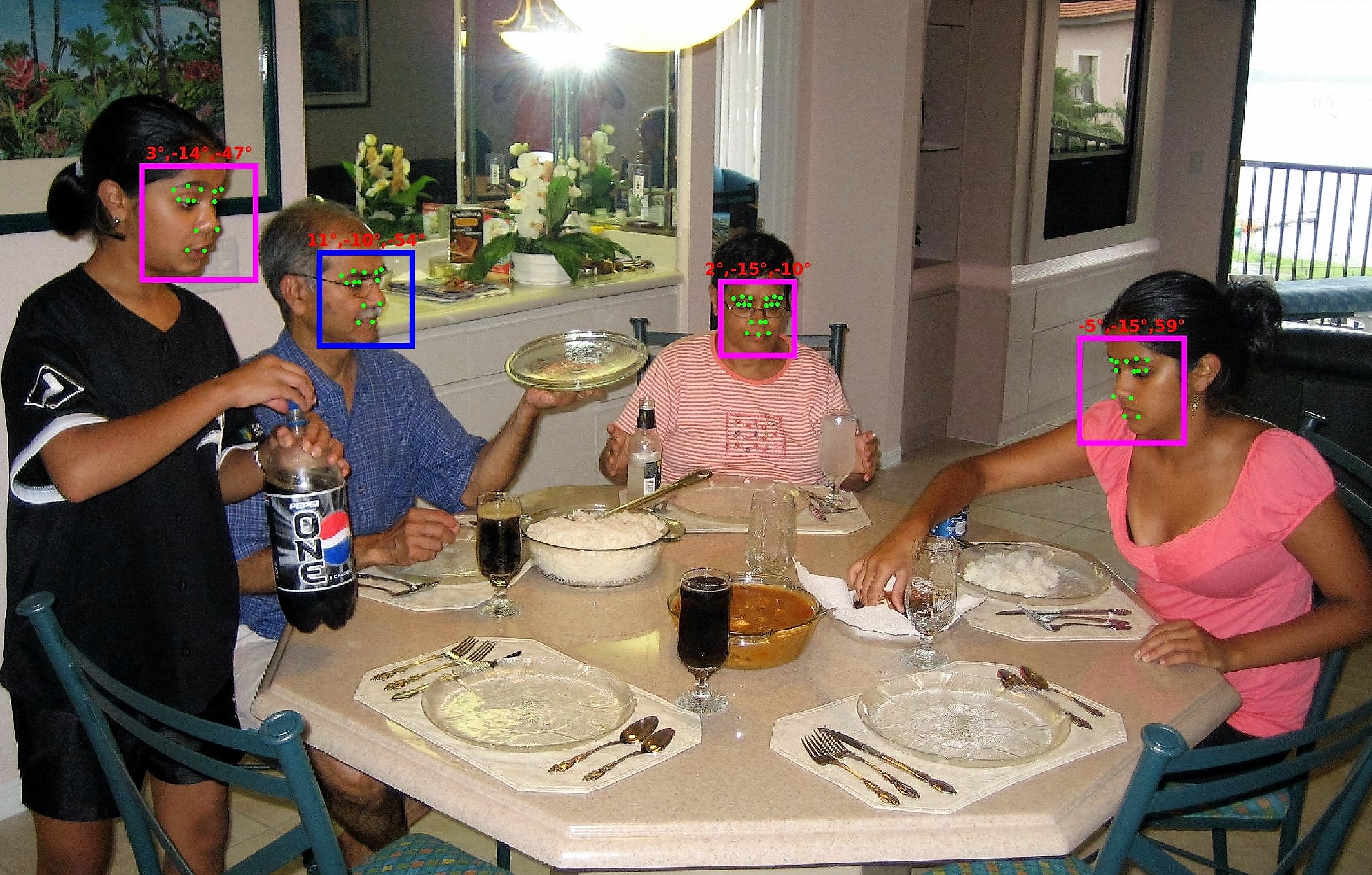}\\
\vskip5pt
\includegraphics[width=5.5cm, height=3.85cm]{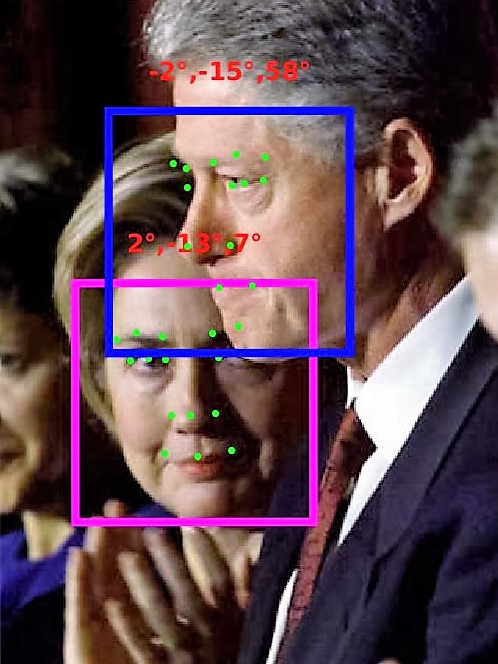}\hskip5pt\includegraphics[width=5.5cm,height=3.85cm]{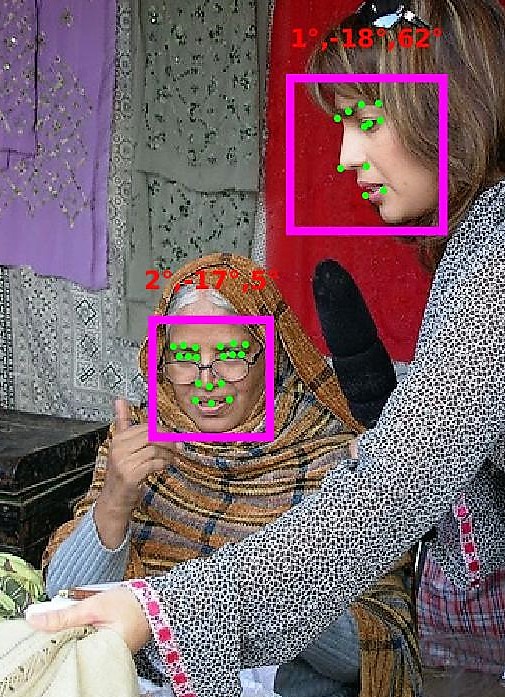}\hskip5pt\includegraphics[width=5.5cm,height=3.85cm]{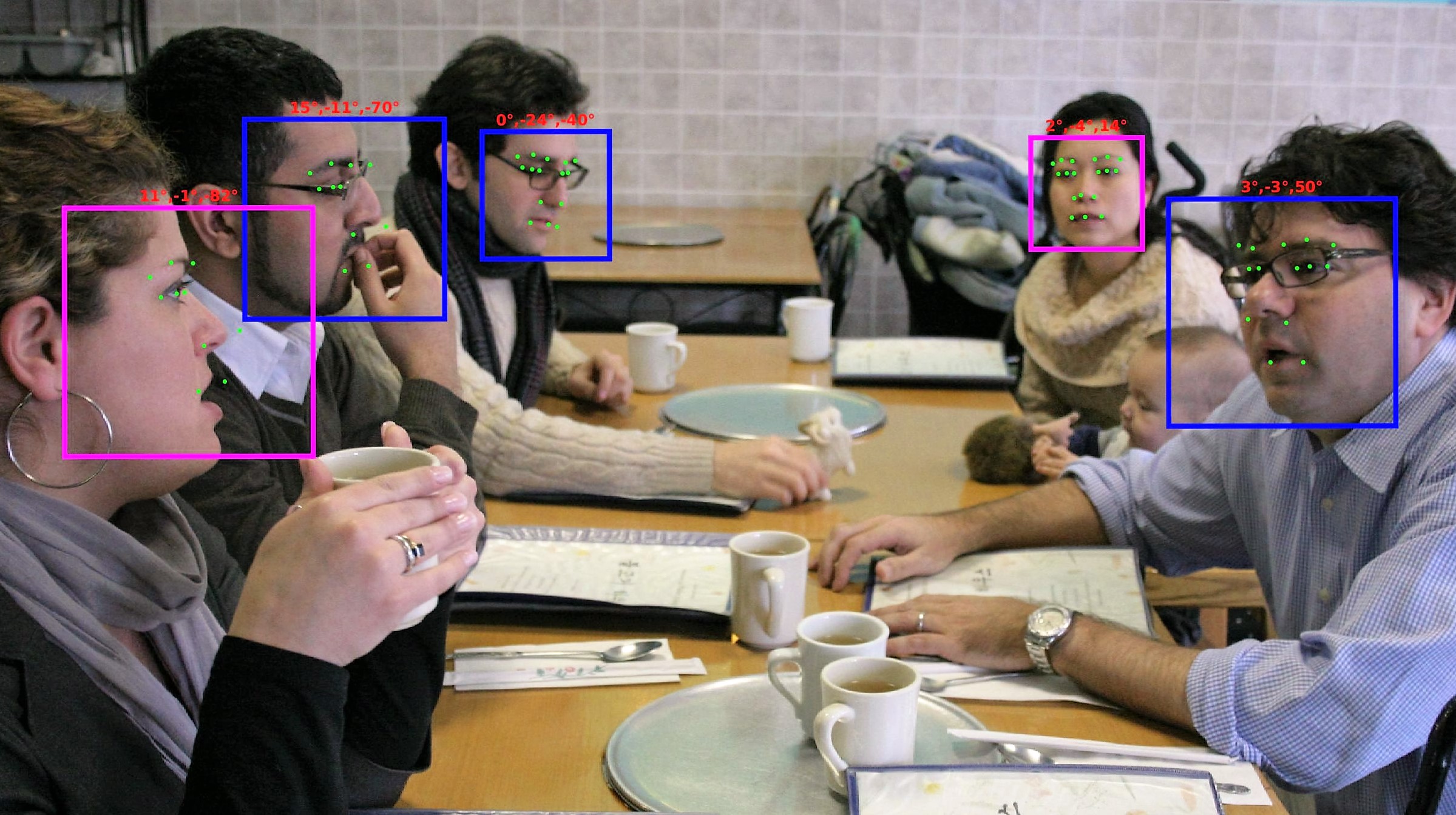}\\
\vskip5pt
\includegraphics[width=5.5cm, height=3.85cm]{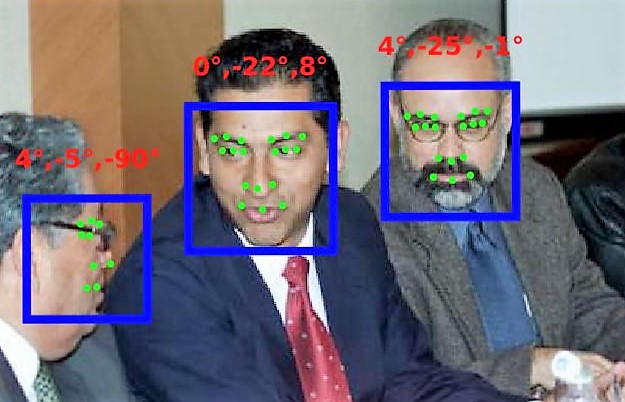}\hskip5pt\includegraphics[width=5.5cm,height=3.85cm]{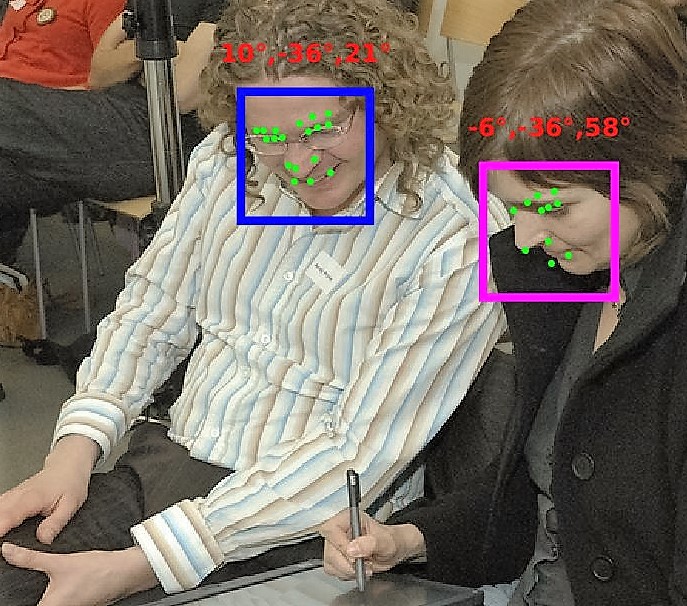}\hskip5pt\includegraphics[width=5.5cm,height=3.85cm]{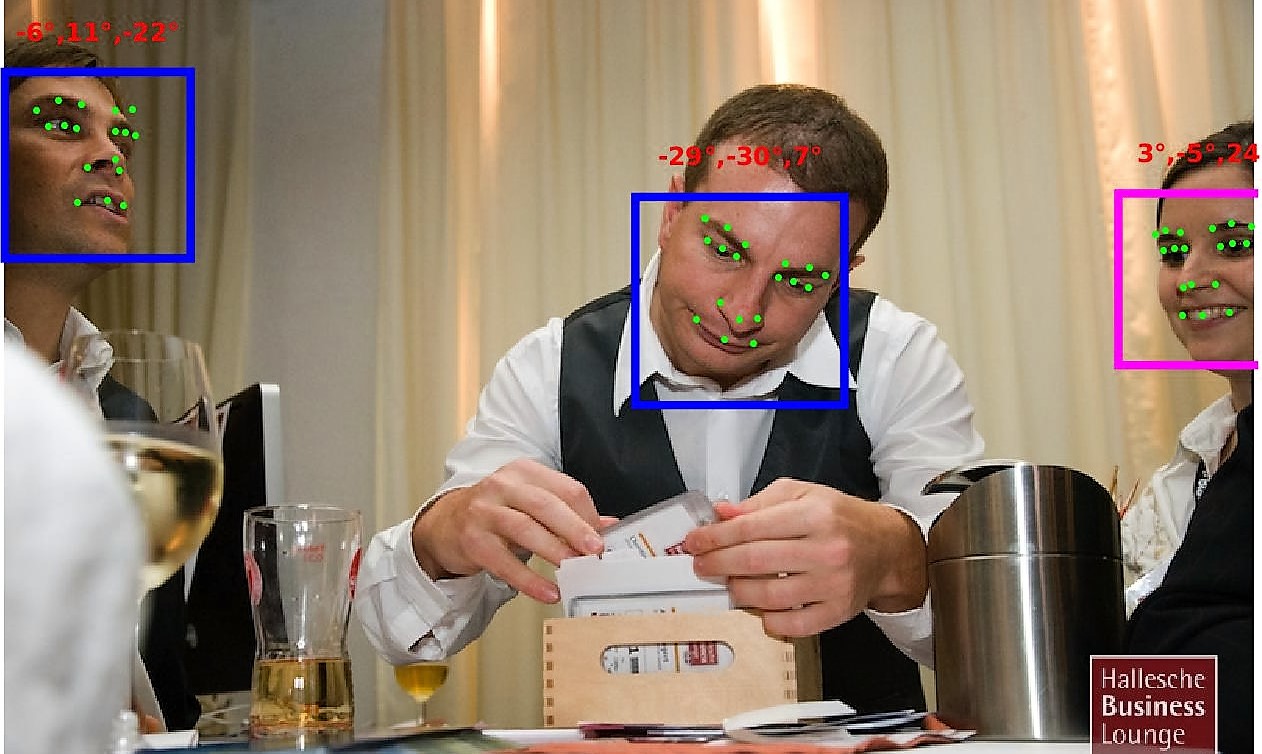}\\
\vskip5pt
\includegraphics[width=5.5cm, height=3.85cm]{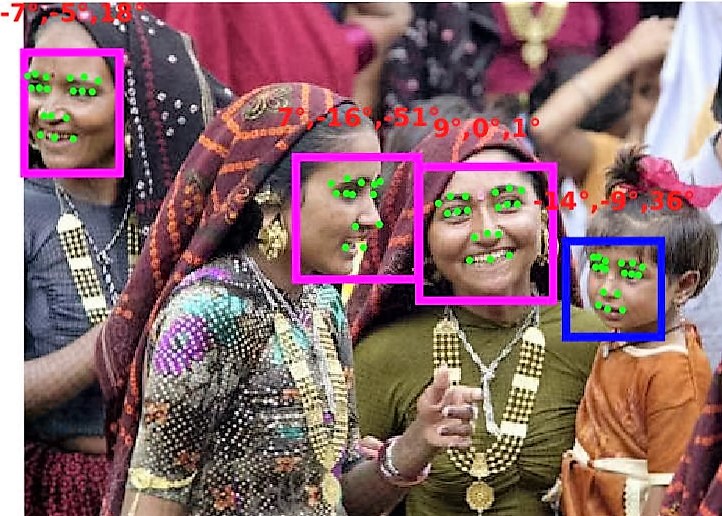}\hskip5pt\includegraphics[width=5.5cm,height=3.85cm]{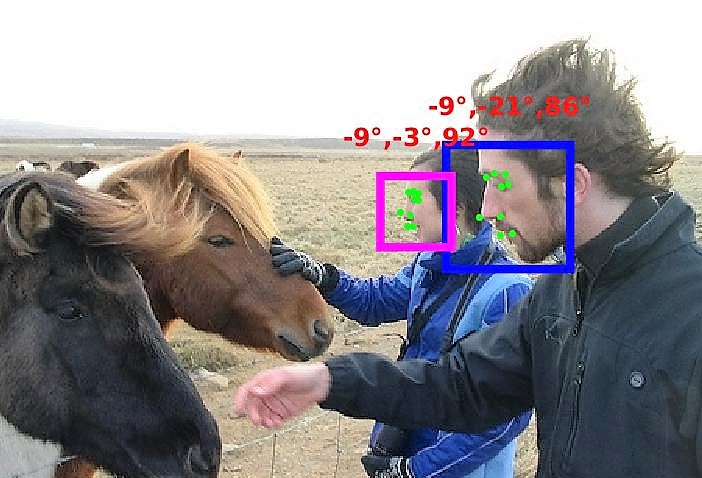}\hskip5pt\includegraphics[width=5.5cm,height=3.85cm]{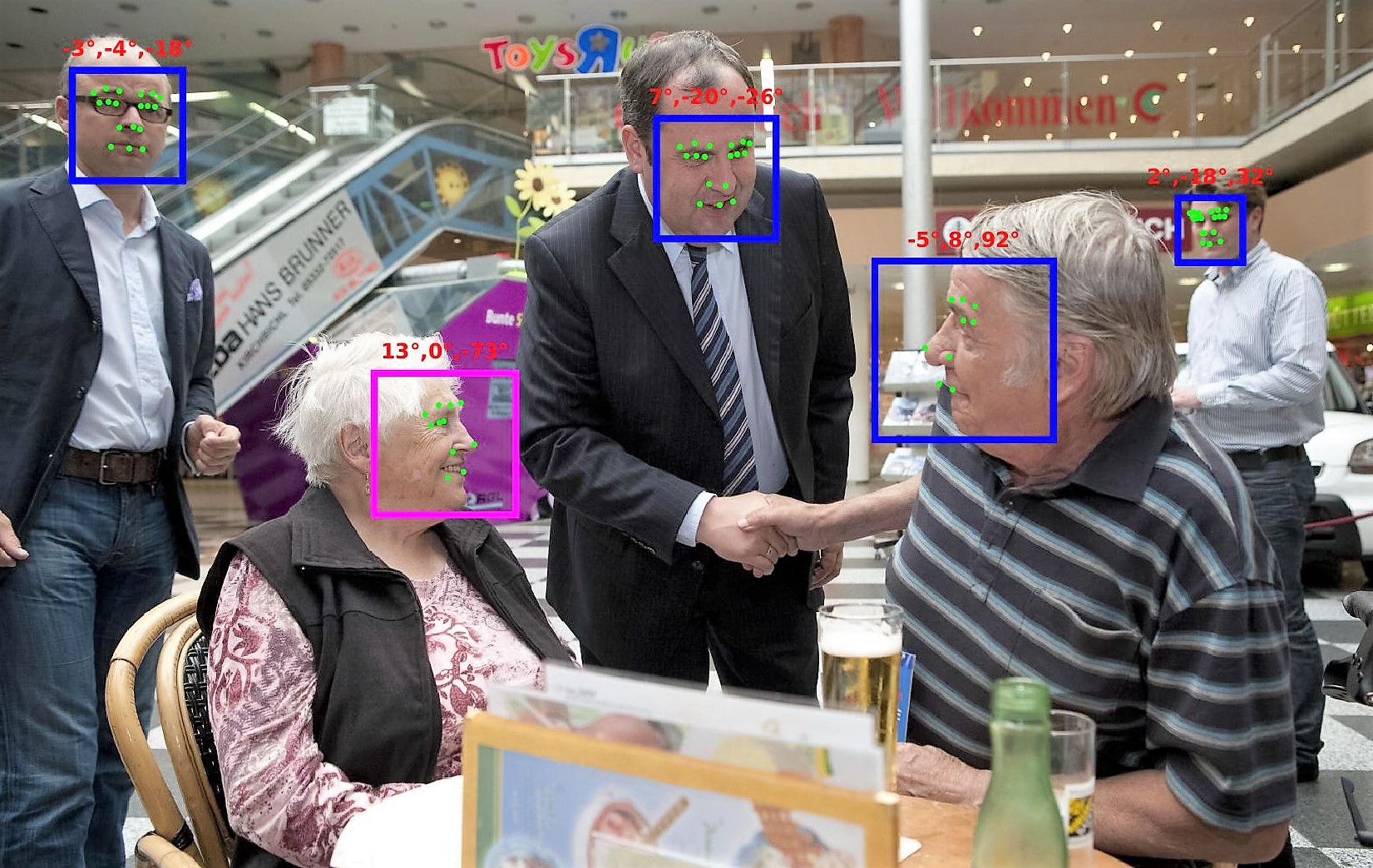}\\
\vskip5pt
\caption{Qualitative results of our method.  The blue boxes denote detected male faces, while pink boxes denote female faces. The green dots provide the landmark locations. Pose estimates for each face are shown on top of the boxes in the order of roll, pitch and yaw.}
\label{fig:quantative_results}
\end{figure*}

\subsection{Fast-HyperFace}
\label{sec:fast-hyperface}
The Hyperface method is tested on a machine with 8 cores and GTX TITAN-X GPU. The overall time taken to perform all the four tasks is $3s$ per image. The limitation is not because of CNN, but due to Selective Search~\cite{vandeSande:2011:SSS:2355573.2356474} algorithm which takes approximately $2s$ to generate candidate region proposals. One forward pass through the HyperFace network for $200$ proposals takes merely $0.1$s.

We also propose a fast version of HyperFace which uses a high recall fast face detector instead of Selective Search~\cite{vandeSande:2011:SSS:2355573.2356474} to generate candidate region proposals. We implement a face detector using Single Shot Detector~(SSD)~\cite{liu2016ssd} framework. The SSD-based face detector takes a $512 \times 512$ dimensional input image and generates face boxes in less than $0.05$s, with confidence probability scores ranging from $0$ to $1$. We use a probability threshold of $0.01$ to select high recall detection boxes. Unlike traditional SSD, we do not use non-maximum suppression on the detector output, so that we have more number of region proposals. Typically, the SSD face detector generates $200$ proposals per image. These proposals are directly passed through HyperFace to generate face detection scores, localize face landmarks, estimate pose and recognize gender for every face in the image. Fast-HyperFace consumes a total time of $0.15$s ($0.05$s for SSD face detector, and $0.1$s for HyperFace) on a GTX TITAN X GPU. The Fast-HyperFace achieves a mAP of $97.6\%$ on AFW face detection task, which is comparable to the HyperFace mAP of $97.9\%$. Thus, Fast-HyperFace improves the speed by a factor of $12$ with negligible degradation in performance.

\hfill

\hfill

\section{Discussion}
\label{sec:discussion}
We present some observations based on our experiments. First, all the face related tasks benefit from using the multi-task learning framework. The gain is mainly due to the network's ability to learn more discriminative features, and post-processing methods which can be leveraged by having landmarks as well as detection scores for a region. Secondly, fusing intermediate layers improves the performance for structure dependent tasks of pose estimation and landmarks localization, as the features become  invariant to geometry in deeper layers of CNN. The HyperFace exploits these observations to improve the performance for all the four tasks. 

We also visualize the features learned by the HyperFace network. Figure~\ref{fig:vizualization} shows the network activation for a few selected feature maps out of $192$ from the $conv_{all}$ layer. It can be seen that some feature maps are dedicated solely for a single task while others can be used to predict different tasks. For example, feature map $27$ and $186$ can be used for face detection and gender recognition, respectively. The former distinguishes the face and non-face regions whereas the latter outputs high activation for the female faces.  Similarly, feature map $19$ shows high activation near eyes and mouth regions, while  feature map $96$ gives a rough contour of the face orientation. These features can be used for landmark localization and pose estimation tasks.

Several qualitative results of our method on the AFW, PASCAL and FDDB datasets are shown in Figure~\ref{fig:quantative_results}.  As can be seen from this figure, our method is able to simultaneously perform all the four tasks on images containing extreme pose, illumination, and resolution variations with cluttered background.

\section{Conclusion}
\label{sec:conc}
In this paper, we presented a multi-task deep learning method called HyperFace for simultaneously detecting faces, localizing landmarks, estimating head pose and identifying gender.  Extensive experiments using various publicly available unconstrained datasets demonstrate the effectiveness of our method on all four tasks. In future, we will evaluate the performance of our method on other applications such as simultaneous human detection and human pose estimation, object recognition and pedestrian detection.

\section*{Acknowledgments}
We thank Dr. Jun-Cheng Chen for implementing the SSD512-based face detector used in the Fast-HyperFace pipeline. 
This research is based upon work supported by the Office of the Director of National Intelligence (ODNI), Intelligence Advanced Research Projects
Activity (IARPA), via IARPA R\&D Contract No. 2014-14071600012. The views and conclusions contained herein are those of the authors and should
not be interpreted as necessarily representing the official policies or endorsements, either expressed or implied, of the ODNI, IARPA, or the U.S. Government. The U.S. Government is authorized to reproduce and distribute reprints for Governmental purposes notwithstanding any copyright annotation
thereon.

{\small
\bibliographystyle{ieee}
\bibliography{HyperFace}
}

\begin{IEEEbiography}[{\includegraphics[width=1in,height=1.50in,clip,keepaspectratio]{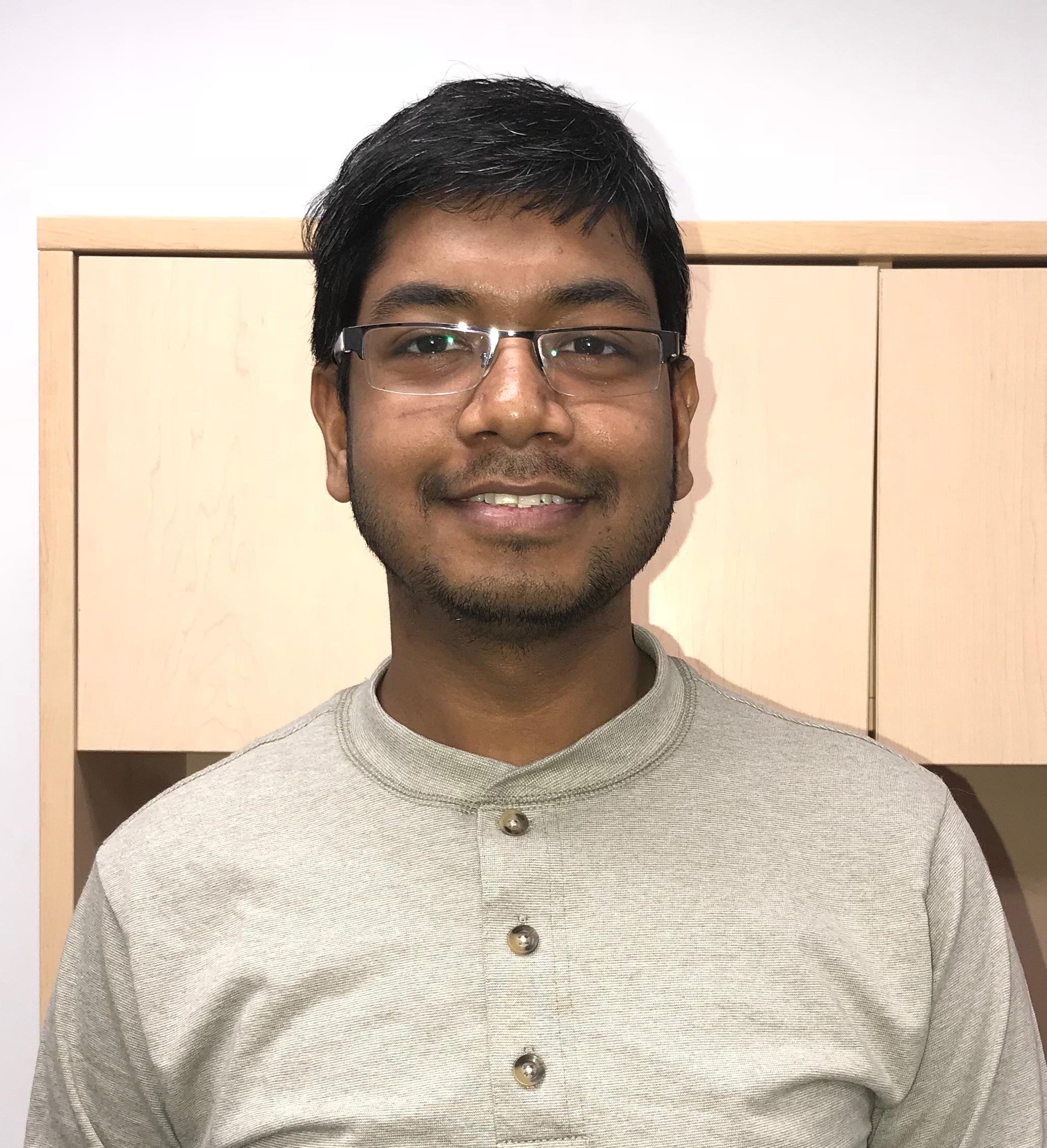}}]{Rajeev~Ranjan}
received the B.Tech. degree in Electronics and Electrical Communication Engineering from Indian Institute of Technology Kharagpur, India, in 2012. He is currently a Research Assistant at University of Maryland College Park. His research interests include face detection, face recognition and machine learning. He received Best Poster Award at IEEE BTAS 2015. He is a recipient of UMD Outstanding Invention of the Year award 2015, in the area of Information Science. He received the 2016 Jimmy Lin Award for Invention.
\end{IEEEbiography}
\begin{IEEEbiography}[{\includegraphics[width=1in,height=1.25in,clip,keepaspectratio]{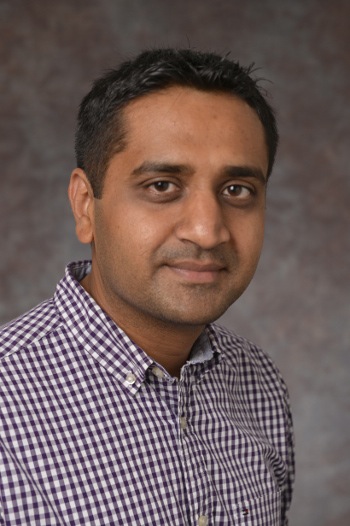}}]{Vishal M. Patel}
received the B.S. degrees in electrical engineering and applied mathematics (Hons.) and the M.S. degree in applied mathematics from North Carolina State University, Raleigh, NC, USA, in 2004 and 2005, respectively, and the Ph.D. degree in electrical engineering from the University of Maryland College Park, MD, USA, in 2010. He is currently an A. Walter Tyson Assistant Professor in the Department of Electrical and Computer Engineering (ECE) at Rutgers University.  Prior to joining Rutgers University, he was a member of the research faculty at the University of Maryland Institute for Advanced Computer Studies (UMIACS). His current research interests include signal processing, computer vision, and pattern recognition with applications in biometrics and imaging. He has received a number of awards including the 2016 ONR Young Investigator Award, the 2016 Jimmy Lin Award for Invention,  A. Walter Tyson Assistant Professorship Award, the Best Paper Award at IEEE BTAS 2015, and Best Poster Awards at BTAS 2015 and 2016.  He is an Associate Editor of the IEEE Signal Processing Magazine,  IEEE Biometrics Compendium, and serves on the Information Forensics and Security Technical Committee of the IEEE Signal Processing Society.   He is a member of Eta Kappa Nu, Pi Mu Epsilon, and Phi Beta Kappa.\end{IEEEbiography}
\begin{IEEEbiography}[{\includegraphics[width=1in,height=1.25in,clip,keepaspectratio]{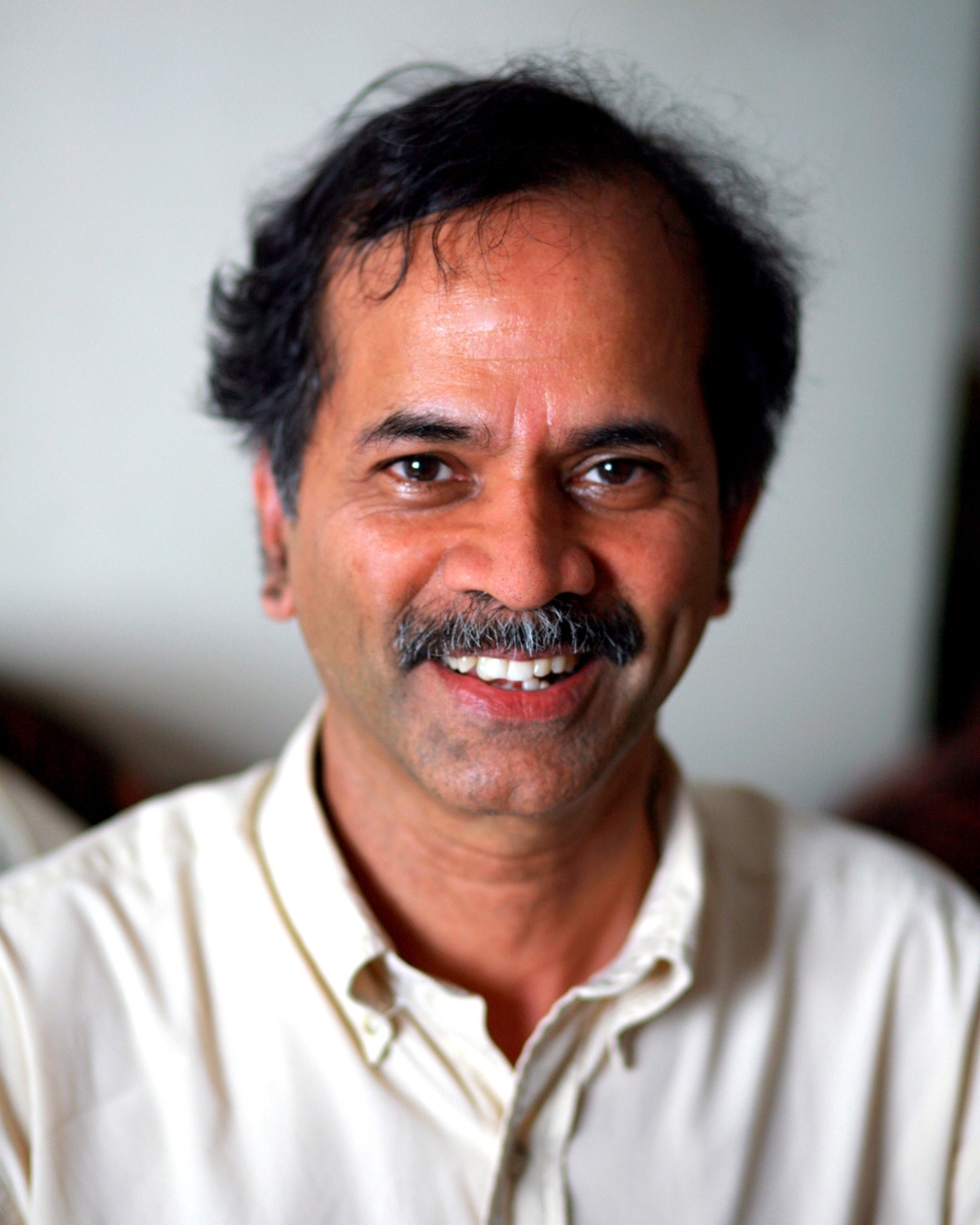}}]{Rama Chellappa}
is a Distinguished University Professor, a Minta Martin Professor of Engineering and Chair of the ECE department at the University of Maryland. Prof. Chellappa received the K.S. Fu Prize from the International Association of Pattern Recognition (IAPR). He is a recipient of the Society, Technical Achievement and Meritorious Service Awards from the IEEE Signal Processing Society and four IBM faculty Development Awards. He also received the Technical Achievement and Meritorious Service Awards from the IEEE Computer Society. At UMD, he received college and university level recognitions for research, teaching, innovation and mentoring of undergraduate students. In 2010, he was recognized as an Outstanding ECE by Purdue University. Prof. Chellappa served as the Editor-in-Chief of PAMI. He is a Golden Core Member of the IEEE Computer Society, served as a Distinguished Lecturer of the IEEE Signal Processing Society and as the President of IEEE Biometrics Council. He is a Fellow of IEEE, IAPR, OSA, AAAS, ACM and AAAI and holds four patents.
\end{IEEEbiography}

\end{document}